\documentclass[conference]{IEEEtran}
\IEEEoverridecommandlockouts
\usepackage{cite}
\usepackage{amsmath,amssymb,amsfonts}
\usepackage{algorithmic}
\usepackage{graphicx}
\usepackage{textcomp}
\usepackage{xcolor}

\usepackage[utf8]{inputenc} 
\usepackage[T1]{fontenc}    
\usepackage{hyperref}       
\usepackage{url}            
\usepackage{booktabs}       
\usepackage{nicefrac}       
\usepackage{microtype}      
\usepackage{enumitem}
\usepackage{xspace}

\usepackage{multirow}

\ifCLASSOPTIONcompsoc
\usepackage[caption=false,font=normalsize,labelfon
t=sf,textfont=sf]{subfig}
\else
\usepackage[caption=false,font=footnotesize]{subfi
g}
\fi

\usepackage{xcolor} 
\newcommand{\rev}[1]{#1}

\def\BibTeX{{\rm B\kern-.05em{\sc i\kern-.025em b}\kern-.08em
    T\kern-.1667em\lower.7ex\hbox{E}\kern-.125emX}}

\iffalse
    \newcommand{\hk}[1]{\textcolor{orange}{[Harsha: #1 ]}}
    \newcommand{\xsq}[1]{\textcolor{purple}{[Shangqing: #1]}}
    \newcommand{\aditya}[1]{\textcolor{red}{[Aditya: #1 ]}}
    \newcommand{\changed}[1]{\textcolor{blue}{[#1]}}
\else
    \newcommand{\hk}[1]{}
    \newcommand{\xsq}[1]{}
    \newcommand{\aditya}[1]{}
    \newcommand{\changed}[1]{#1}
\fi

\newcommand{\model}{\textsc{HiereInterpret}\xspace}

\begin{document}

\title{Hierarchical Industrial Demand Forecasting with Temporal and Uncertainty Explanations
}

\author{
\IEEEauthorblockN{Harshavardhan Kamarthi\textsuperscript{\textsection}}
\IEEEauthorblockA{\textit{Georgia Institute of Technology} \\
Atlanta, USA\\
hkamarthi3@gatech.edu}
\and
\IEEEauthorblockN{Shangqing Xu\textsuperscript{\textsection}}
\IEEEauthorblockA{\textit{Georgia Institute of Technology} \\
Atlanta, USA\\
sxu452@gatech.edu}
\and
\IEEEauthorblockN{Xinjie Tong\textsuperscript{*}}
\IEEEauthorblockA{\textit{Aspen Technology} \\
Houston, USA \\
xinjietong001@gmail.com}
\and
\IEEEauthorblockN{Xingyu Zhou}
\IEEEauthorblockA{\textit{The Dow Chemical Company} \\
Houston, USA \\
xzhou14@dow.com}
\and
\IEEEauthorblockN{James Peters}
\IEEEauthorblockA{\textit{The Dow Chemical Company} \\
Midland, USA \\
japeters@dow.com}
\and
\IEEEauthorblockN{Joseph Czyzyk}
\IEEEauthorblockA{\textit{The Dow Chemical Company} \\
Midland, USA \\
jjczyzyk@dow.com}
\and
\IEEEauthorblockN{B. Aditya Prakash}
\IEEEauthorblockA{\textit{Georgia Institute of Technology} \\
Atlanta, USA\\
badityap@cc.gatech.edu}
}

\maketitle

\begingroup\renewcommand\thefootnote{\textsection}
\footnotetext{Equal contribution}
\endgroup
\begingroup\renewcommand\thefootnote{*}
\footnotetext{The author was with The Dow Chemical Company, Houston, USA, when this work was performed.}
\endgroup

\begin{abstract}
Hierarchical time-series forecasting is essential for demand prediction across various industries.
While machine learning models have obtained significant accuracy and scalability \rev{on such forecasting tasks}, the interpretability of their predictions, informed by application, is still largely unexplored.
To bridge this gap, we introduce a novel interpretability method for large hierarchical probabilistic time-series forecasting, adapting generic interpretability techniques while addressing challenges associated with hierarchical structures and uncertainty.
Our approach offers valuable interpretative insights in response to real-world industrial supply chain scenarios, including 1) the significance of various time-series within the hierarchy and external variables at specific time points, 2) the impact of different variables on forecast uncertainty, and 3) explanations for forecast changes in response to modifications in the training dataset. 
To evaluate the explainability method, we generate semi-synthetic datasets based on real-world scenarios of explaining hierarchical demands for over ten thousand products at a large chemical company. The experiments showed that our explainability method successfully explained state-of-the-art industrial forecasting methods with significantly higher explainability accuracy, both in point and probabilistic forecasts, across a wide range of benchmarks with good scalability.
Furthermore, we provide multiple real-world case studies that show the efficacy of our approach in identifying important patterns and explanations that help various stakeholders better understand the forecasts. 
Additionally, our method facilitates the identification of key drivers behind forecasted demand, enabling more informed decision-making and strategic planning. By offering a transparent view of the forecasting process, our approach helps build trust and confidence among users, ultimately leading to better adoption and utilization of hierarchical forecasting models in practice.
\end{abstract}


\section{Introduction}

\subsection{\changed{Background}}
Hierarchical time-series forecasting (HTSF) \cite{fliedner2001hierarchical} provides crucial insights for many domains, including industrial demand forecasting. 
For example, for a large manufacturing company, forecasting future demand aggregated across different products and geographical levels is the foundation for production planning, fulfillment, and inventory management \cite{bose2017probabilistic}. 
Accurate forecasts at different levels of the hierarchy enable better decision-making and resource allocation, ultimately leading to cost savings and improved efficiency. 
Moreover, HTSF models can capture the inherent structure of the data, allowing for more granular insights and the ability to drill down into specific segments or regions. 
This hierarchical approach not only improves the accuracy of the forecasts but also provides a more comprehensive understanding of the underlying patterns and trends. 
By leveraging HTSF, companies can better anticipate market demands, optimize their supply chains, and respond more effectively to changes in the market environment.

State-of-the-art HTSF models treat input series as samples from probabilistic distributions and output estimated time-series distributions \cite{salinas2019high, rangapuram2021end, han2021simultaneously, kamarthi2023rigidity}. 
These models leverage advanced deep learning techniques to capture complex temporal patterns and dependencies across different hierarchical levels. 
Aside from such advancements on general HTSF datasets, some have also investigated how to build HTSF models that are specifically designed to adapt to large-scale industrial demands \cite{kamarthi2024large}. 
These specialized models address the unique challenges posed by industrial datasets, such as handling large volumes of data, dealing with missing values, and incorporating domain-specific knowledge into the forecasting process. 
Moreover, they often include mechanisms to ensure scalability and robustness, which are critical for practical deployment in industrial settings. 
By integrating these specialized techniques, HTSF models can provide more accurate and reliable forecasts, ultimately supporting better decision-making and operational efficiency in industrial applications.

\begin{figure}[ht]
    \centering
    \includegraphics[width=0.99\linewidth]{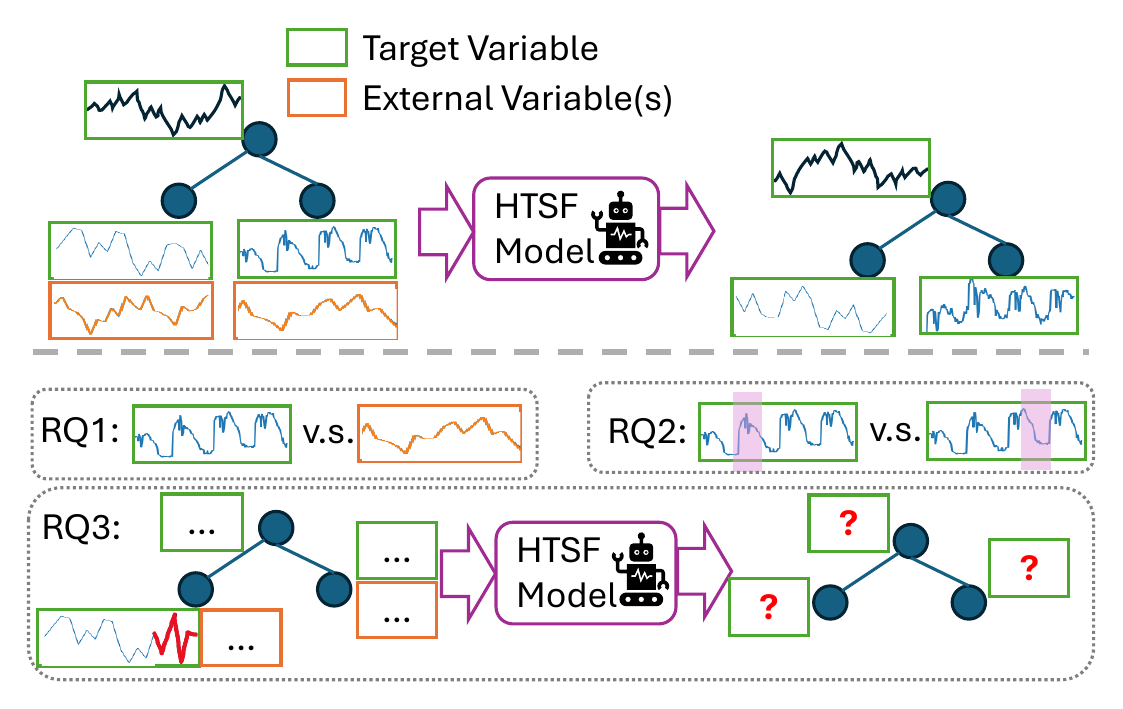}
    \caption{\changed{In real-world HTSF tasks, stakeholders and planners want to know: RQ1: which input variable(s) contribute most to the prediction; RQ2: which time steps of each variable contribute most to the prediction; RQ3: why prediction results change when input data changes.}}
    \label{fig:introFig}
\end{figure}

Yet, state-of-the-art HTSF models are deep-learning models that are black boxes, which fall short of providing explanations for their predictions. 
Interpretability is crucial for many real-world applications, especially in industrial demand forecasting, where stakeholders need to understand the underlying factors driving the forecasts.
For example, they may want to know which variables are most important for the predictions, which time steps in the input histories are most critical, and why the prediction results change when there are changes in the input data. \rev{Such industrial forecasting scenarios are mostly coupled with large-scale data that may contain complex hierarchical structures, imposing unique challenges from underlying data engineering efforts.}
However, existing interpretability methods for time-series forecasting are not directly applicable to HTSF tasks, particularly in industrial demand forecasting.
These methods are not designed to handle the complex hierarchical structure of industrial demand data, which contains a large number of time-series with underlying relationships between them.
This drawback significantly affects industrial demand forecasting and blocks their deployment into real-world use cases.
For example, as shown in Fig. \ref{fig:introFig}, stakeholders and planners want to know:
\begin{enumerate}
    \item \textit{RQ1:} Which variables contribute the most to the predictions: Understanding the key drivers behind the forecasted values is essential for decision-making. Identifying the most influential variables helps in focusing on the critical factors that impact demand, allowing for more targeted interventions and strategies.
    \item \textit{RQ2:} Which time steps in input histories are most important: Knowing which specific time periods in the historical data are most relevant to the predictions can provide insights into temporal patterns and trends. This information is valuable for understanding how past events influence future demand.
    \item \textit{RQ3:} Why prediction results change when there are changes in the input data: It is crucial to understand the sensitivity of the model to changes in the input variables. This helps in assessing the robustness of the forecasts and in identifying potential risks or opportunities associated with changes in the market environment or other external factors.
\end{enumerate}
Such questions cannot be answered by existing models.

Although multiple interpretability methods have been developed to explain machine-learning black-box models, including general time-series forecasting \cite{ribeiro2016should, zeiler2014visualizing, sundararajan2017axiomatic, selvaraju2017grad, smilkov2017smoothgrad}, they cannot be directly adapted to HTSF tasks, especially industrial demand forecasting tasks. First, industrial demands contain an extremely large number of time-series that have underlying relationships between them due
to hierarchical constraints.
Naively running existing interpretability methods on such a deep hierarchical structure
would provide suboptimal results.
This is because these methods cannot handle such high-dimensional inputs that
have both complex temporal and hierarchical relations.
Further, applying these methods to such high-dimensional multivariate hierarchical
time-series imposes significant computational costs.
Further, existing interpretability methods are designed for models with deterministic outputs, while state-of-the-art HTSF models all output probabilistic distributions. 
Adapting them to different kinds of probabilistic hierarchical forecasting models
with different distributional assumptions
is non-trivial.
Finally, there's no public time-series dataset with ground truth explanations, meaning it's hard to evaluate. 

\subsection{\changed{Our Contribution}}

To address the challenges, we propose \changed{two novel modifications} to adapt interpretability methods for general time-series tasks to HTSF. First, we propose a subtree approximation modification, which decomposes cross-hierarchy importance into adjacent-hierarchy importance. This approach simplifies the computational complexity while ensuring that the explanation is aligned with hierarchical coherency. By breaking down the hierarchical structure into smaller, more manageable subtrees, we can more effectively capture the relationships and dependencies within the data, leading to more accurate and interpretable results. Second, we propose to find a deterministic alternative for probabilistic models with any unknown distribution by using quantiles of the output distribution as alternative outputs. This method allows us to approximate the behavior of probabilistic models in a deterministic manner, making it easier to apply existing interpretability techniques. By leveraging quantiles, we can capture the key characteristics of the probabilistic forecasts, providing a more intuitive and understandable explanation of the model's predictions. 

These two modifications enable us to adapt a wide range of general interpretability methods to HTSF tasks, addressing the unique challenges posed by the hierarchical structure and probabilistic nature of industrial demand data. Our approach not only improves the interpretability of HTSF models but also enhances their practical applicability in real-world industrial settings.

\changed{Furthermore, to overcome the lack of \rev{datasets and evaluate} how powerful our proposed method can be in explaining real-world industrial prediction scenarios}, we propose to build a synthetic benchmark with both known explanation and real-world distribution.
We achieve this by first building hierarchical synthetic series with different anomaly patterns and dependencies, then adding synthetic series to real-world datasets.
To perform a sound evaluation, we assign such anomalies to different levels of hierarchy as well as different variables. 

We evaluated the methods on the established benchmark regarding both deterministic and probabilistic settings.
In the deterministic setting, our subtree approximation brings an average $62.0\%$ improvement when calculating all-variable dependencies and $12.3\%$ improvement when calculating target-variable-only dependencies. 
In the probabilistic setting, we show that our method adapts a wide range of probabilistic forecasting methods. We observe 10-25\% improvement due to subtree approximation.

In addition to synthetic results, we also conducted case studies on original real-world data from Dow's demand data. Results show our method explains how an HTSF model treats the historical trends as well as the relationship between external variables well. 
\changed{Also, we show case studies on real-world industrial demand forecasting scenarios} where our model helps understand the changes in forecasts while part of the input variable changes, which is an important topic verified by Dow's domain expert. 

Our contribution can be summarized into the following points:
\begin{enumerate}
    \item \changed{\textit{Proposed Solution}:} We are the first to propose a generalized solution for adapting a wide range of general explainability methods for hierarchical and probabilistic forecasting tasks. Our approach addresses the unique challenges posed by the hierarchical structure and probabilistic nature of industrial demand data, providing more accurate and interpretable forecasts.
    \item \changed{\textit{Established Benchmark}:} We established a new benchmark that evaluates models in identifying different kinds of explanations and leverages real-world datasets to generate these semi-synthetic benchmarks. This benchmark includes various anomaly patterns and dependencies, allowing for a comprehensive evaluation of interpretability methods in hierarchical time-series forecasting.
    \item \changed{\textit{Evaluation}:} We evaluated the methods on the synthetic dataset and proved the effectiveness of our proposed adaptations with 12-62\% performance improvement over baseline methods. Our subtree approximation and quantile-based deterministic alternative significantly enhance the interpretability of HTSF models, making them more suitable for practical applications.
    \item \changed{\textit{Real-world Case Studies}:} We conducted case studies and showed our methods have the potential to be adapted to real-world cases, such as identifying important products affecting demand, identifying changes in demand trends due to a pandemic, and explaining changes in variance and uncertainty of forecasts over time due to changes in consumer demand. These case studies demonstrate the practical value of our approach in providing actionable insights for stakeholders.
\end{enumerate}



\section{Related Works}

\paragraph{Hierarchical Probabilistic Forecasting}

\changed{
Traditional hierarchical probabilistic forecasting first forecasts each level of hierarchy independently and then reconciles the forecasts by aggregating across hierarchies \cite{erven2015game, novak2017bayesian, ben2019regularized, corani2020probabilistic, wickramasuriya2024probabilistic}.
More recent hierarchical probabilistic forecasting models are trained end-to-end, producing outputs considering all levels of inputs by approximating hierarchical sequences by Gaussian process \cite{salinas2019high}, re-projecting sampled results to satisfy hierarchical coherency \cite{rangapuram2021end}, tuning nonlinear models by quantile regression loss and applying soft regularization \cite{han2021simultaneously, han2022dynamic}, or introducing graph neural networks to model the hierarchical structure and soft regularization~\cite{sriramulu2024deephgnn, oskarsson2024probabilistic}. Later studies has also proposed to develop hierarchical probabilistic forecasting models for large-scale datasets by soft distributional consistency regularization\cite{kamarthi2023rigidity, kamarthi2024large}. All these methods above are black-box and do not provide explanations for outputs, calling for adapting interpretability metrics.}

\paragraph{Explainable ML}
\changed{
Model interpretability for black-box machine-learning models is an important problem to enable trustworthy and robust machine learning deployments in practice~\cite{doshi2017towards}.
Recent interpretability advancements for black-box machine-learning models focus mainly on post-hoc methods~\cite{bodria2023benchmarking,danilevsky2020survey,zhao2023interpretation}.
Such methods approximate the relationship between the output and the input by model-agnostic measurements~\cite{lakkaraju2019faithful,ribeiro2018anchors} and have shown wide applicability ~\cite{linardatos2020explainable,samek2021explaining}.
Proposed measurements can be categorized into: a) Gradient-based methods \cite{sundararajan2017axiomatic, smilkov2017smoothgrad, selvaraju2017grad}, which measure such relationships based on model gradients regarding the importance of each input feature to the model's output; b) Perturbation-based methods \cite{zeiler2014visualizing, suresh2017clinical,lundberg2017unified, ismail2020benchmarking}, which quantify the output changes after putting a local perturbation on input; c) Approximation-based methods \cite{castro2009polynomial, ribeiro2016should, shrikumar2017learning}, which distill the black-box model to a white-box model and then get corresponding explanations.}

\paragraph{Interpretability for Time-Series Models}

\changed{The adaptation of black-box interpretability methods into time-series tasks is also investigated. The mainstream investigations reprogram interpretability scores under time-series tasks' input-output format by temporal rescaling \cite{garreau2020explaining} or temporal importance awareness \cite{tonekaboni2020went}. Some of the efforts have also evaluated general interpretability methods on synthetic benchmarks \cite{gandin2021interpretability, turbe2023evaluation}, as most public time-series databases are not coupled with ground-truth explanations for evaluation.
Yet, to the best of our knowledge, existing work has not adapted interpretability methods into hierarchical and probabilistic forecasting, which is an important challenge in the real-world deployment of demand forecasting models.} \rev{Furthermore, applying the interpretability methods to large-scale databases imposes unique challenges, as verified by efforts in other domains such as cloud-based management \cite{yan2025east}, neural topic modeling \cite{gao2024enhancing}, and federated client estimation \cite{wang2024fast}, which is still underexplored in industrial demand forecasting scenarios.}


\section{Preliminary}

\paragraph{Hierarchical Probabilistic Forecasting}

Hierarchical time-series forecasting is a special case of multivariate forecasting where the underlying time-series have hierarchical relations between them.
Formally, we note a hierarchical dataset $\mathcal{D} = \{x_1, \dots, x_N\}$ as a dataset consisting of $N$ time-series. 
Each series $x_i = \{x_i^1, \dots, x_i^T\}$ is composed of $T$ time steps following an unknown distribution $\mathcal{P}_i$. 
Each time step $x_i^t$ contains one target variable that we aim to predict and multiple external variables that are used as assistance for prediction. 
Further, the target variable of all series satisfies a hierarchical relationship $\tau = (G_\tau, H_\tau)$, where $G_\tau$ is a tree-form dependency map and $H_\tau = \{x_i = \sum_{(i,j) \in G_\tau } \phi_{ij}x_j\}$ is the restraint across hierarchies parameterized by a known factor $\phi_{ij}$.

A hierarchical probabilistic forecasting model $f$ takes $\mathcal{D}$ as input and yield predictions $\hat{x_i} = \{\hat{x}_i^{T+1}, \dots, \hat{x}_i^{T+h}\}$ of prediction window $h$ for each $x_i \in \mathcal{D}$. Specifically, the model will approximate $\mathcal{P}_i$ with certain $\hat{\mathcal{P}}_i(\theta)$ parameterized by $\theta$, then optimize $\theta$ by minimizing certain form of loss function given ground truth $x_i$ and $\hat{\mathcal{P}}_i(\theta)$. The prediction $\hat{x_i}$ will then be sampled from $\hat{\mathcal{P}}$.

\paragraph{Interpretability for Time-Series Models}

Given a black-box machine learning model $f: x^{1\dots T} \rightarrow \hat{x}^{T+1 \dots T+h}$ for general time-series tasks with deterministic output $\hat{x}$, an interpretability method $g$ yields an explanation $I = g(x, f(x)) = \{I(x_{ij})\}$ which represents the importance score of each input component $x_{ij}$ towards output $\hat{x} = f(x)$. We aim to adapt such interpretability methods to hierarchical probabilistic forecasting, where the output $\hat{x}^k$ is sampled from a distribution $\hat{P}_k$ instead.

\section{Methodology}

\subsection{Adaptation to Hierarchical Task}
To adapt interpretability methods to hierarchical time-series, a straightforward solution is to flatten the series and treat hierarchical forecasting as multivariate forecasting.
Yet, such an approach has drawbacks.
First, treating sequences as multivariate ones misses hierarchical coherency across levels, while such coherency is important in hierarchical forecasting \cite{hyndman2011optimal}.
Second, such an approach would calculate the importance score between every output variable and every input variable.
When the dependency tree is deep or wide, such a calculation would not only be computationally heavy but also be noisy, as it's hard to measure the effect of perturbation across multiple levels accurately.
Additionally, this method does not leverage the hierarchical structure, which can provide valuable information about the relationships between different time-series.
By ignoring this structure, the interpretability method may fail to capture important dependencies and interactions that are crucial for accurate forecasting.
Furthermore, the computational complexity of calculating importance scores for all possible pairs of input and output variables can be prohibitive, especially for large datasets with many time-series.
This can lead to increased noise and errors in the importance scores, making the interpretability method less reliable and harder to trust.
Therefore, a more sophisticated approach that takes into account the hierarchical structure of the data is necessary to provide accurate and meaningful importance scores for hierarchical time-series forecasting.

\begin{figure}[ht]
    \centering
    \includegraphics[width=0.99\linewidth]{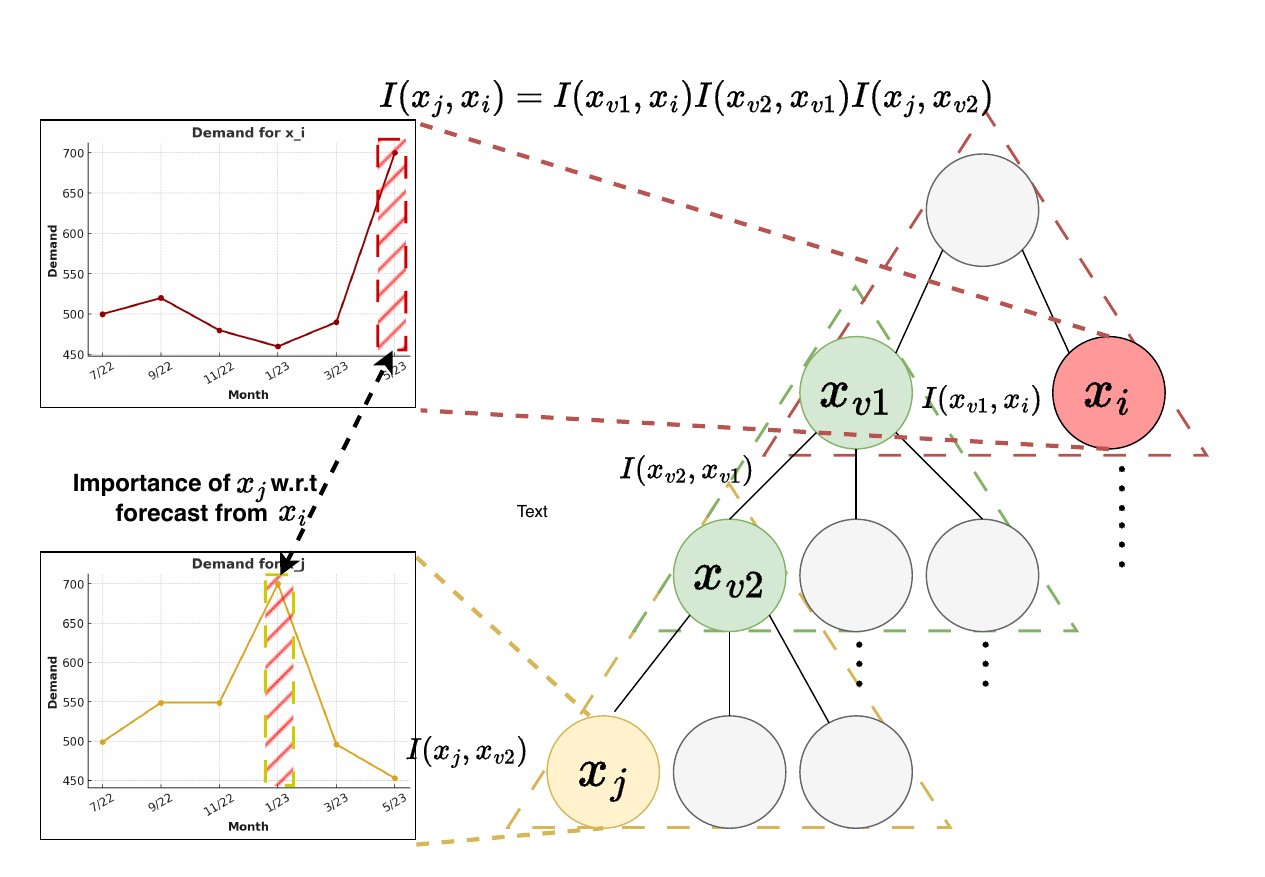}
    \caption{Subtree approximation allows \model to accurately capture importance scores of any time-series across the hierarchy}
    \label{fig:Sub-tree approximation}
\end{figure}

\begin{figure}
    \centering
\begin{algorithmic}[1]
    \REQUIRE Hierarchical structure $\mathcal{H}$, Target node $x_i$, Importance function $I(x, y)$
    \ENSURE Importance scores of all nodes w.r.t. $x_i$

    \STATE Initialize an empty dictionary $\texttt{scores}$ to store importance scores
    \STATE $\texttt{scores}[x_i] \gets 1$
    \STATE Initialize a queue $Q$ and enqueue $(x_i, 1)$
    \STATE Initialize a set $V$ of visited nodes

    \WHILE{$Q$ is not empty}
      \STATE Dequeue $(x_{current}, score_{current})$ from $Q$
      \FORALL{child node $x_{child}$ of $x_{current}$ in $\mathcal{H}$ such that $x_{child} \notin V$}
        \STATE $score_{child} \gets score_{current} \times I(x_{child}, x_{current})$
        \STATE $\texttt{scores}[x_{child}] \gets score_{child}$
        \STATE Enqueue $(x_{child}, score_{child})$ to $Q$
      \ENDFOR
      \STATE Let $x_{parent}$ be the parent of $x_{current}$ in $\mathcal{H}$
      \IF{$x_{parent}$ exists and $x_{parent} \notin V$}
        \STATE $score_{parent} \gets score_{current} \times I(x_{current}, x_{parent})$
        \STATE $\texttt{scores}[x_{parent}] \gets score_{parent}$
        \STATE Enqueue $(x_{parent}, score_{parent})$ to $Q$
      \ENDIF
      \STATE Add $x_{current}$ to $V$
    \ENDWHILE
    \RETURN $\texttt{scores}$
\end{algorithmic}
\caption{Calculate importance scores for forecasting node $x_i$ via subtree approximation}
\label{alg:Hierarchical Importance}
\end{figure}

To address these concerns, we introduce subtree approximation.
While calculating the importance score between series that are not in adjacent hierarchy levels, we assume that such an importance score can be decomposed into chain-form scores between adjacent hierarchies.
Specifically, consider two series $x_i, x_j$ with connection route $x_i \rightarrow x_{v_1} \rightarrow \dots \rightarrow x_{v_m} \rightarrow x_j $ where each $(x_{v_i}, x_{v_{i+1}})$ is in adjacent levels, our assumption can be represented by:
\begin{equation}
    I(x_i, x_j) \approx I(x_i, x_{v_1}) I(x_{v_1}, x_{v_2}) \dots I(x_{v_m},  x_{j})
\end{equation}

By this approximation, we can get the any-to-any importance score by calculating the importance score between adjacent variables $(x_v, x_k),\ (v,k) \in G_{\tau}$ only (Fig. \ref{fig:Sub-tree approximation}, Alg. \ref{alg:Hierarchical Importance}).
This not only helps us reduce the computation costs but also reduces
the noise and errors that arise from dealing with all the time-series simultaneously that are inherently related by hierarchical constraints.
Furthermore, restricting perturbations referring to constraints between adjacent levels would be a lot easier compared to multiple levels.

\subsection{Adaptation to Probabilistic Task} \label{sec:probAdapt}

\begin{figure}[ht]
    \centering
    \includegraphics[width=0.98\linewidth]{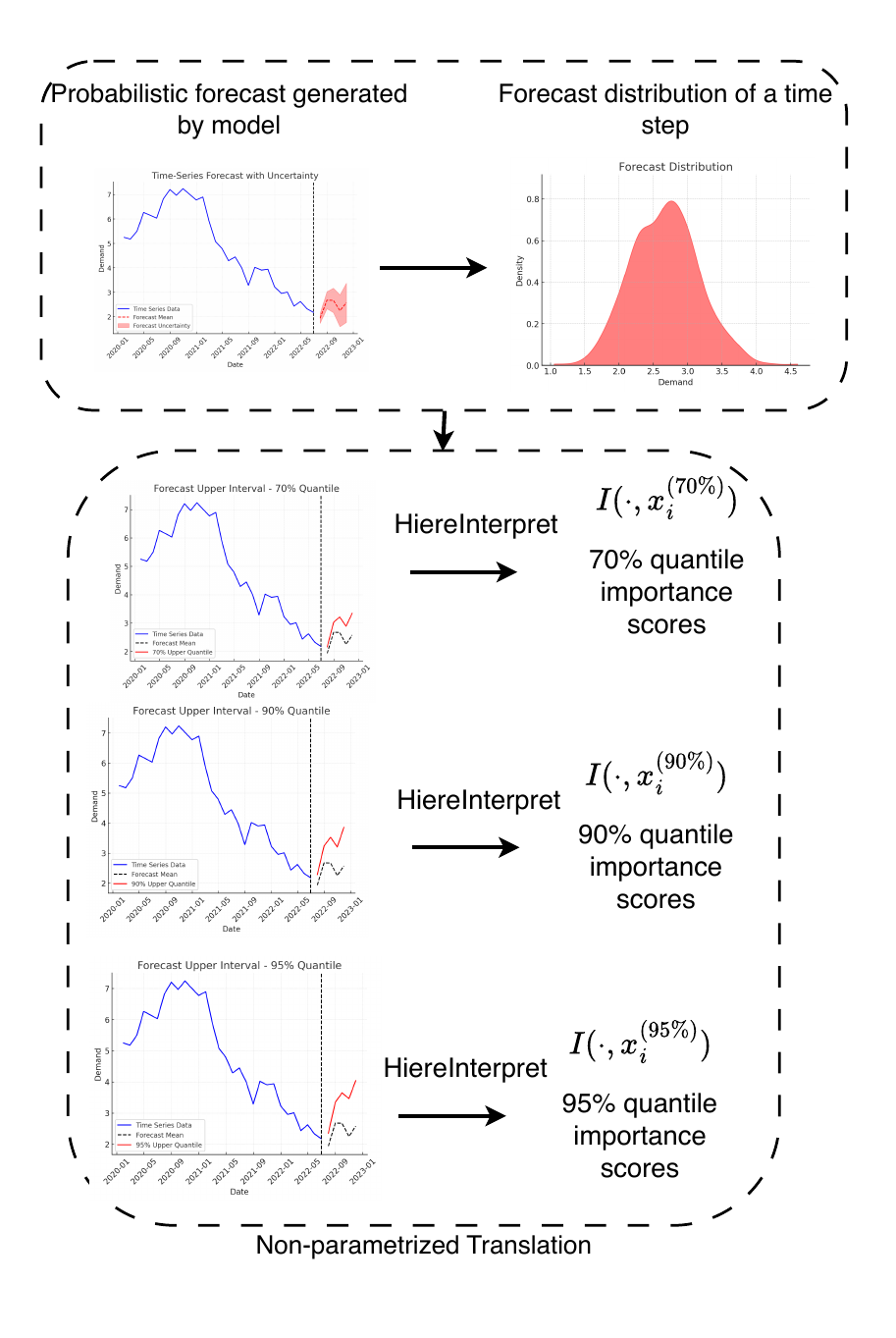}
    \vspace{-0.2in}
    \caption{Adapting to any probabilistic forecast by deriving importance scores from quantiles of the forecasts}
    \label{fig:Probabilistic Adaption}
\end{figure}

To get an unbiased explanation of probabilistic outputs, interpretability methods should calculate an importance score regarding output distributions $\hat{P}$ instead of sampled outputs $\hat{x} \in \hat{P}$.
Solving such a problem is non-trivial, as existing interpretability methods assume the model's output is deterministic.

To tackle the above challenges, we propose to transform probabilistic models into equivalent deterministic models that could be explained by interpretability methods. A straightforward solution is to use the parameters of the output distribution as an output of an alternative model. For example, for \rev{a model $f$ which outputs} a Gaussian distribution $f: x \rightarrow \mathcal{P} = (\mu, \sigma)$, we treat $f': x \rightarrow (\mu, \sigma)$ as an equivalent model and regard the explanation of $f'$ as the explanation of $f$. Yet, this solution does not work for models with an unknown output distribution. Instead, we propose a general \textbf{Non-parameterized Transition} solution that works for \textit{any} probabilistic model. \rev{Specifically, for} a model with unknown output distribution $\hat{P}$, we sample $\tilde{X} = \{\tilde{x}_i\}, \ x_i \in \hat{P}$ from the output distribution and calculate the quantiles $q(\tilde{X})$. This approach allows us to capture the variability and uncertainty inherent in the probabilistic forecasts, providing a more comprehensive understanding of the model's behavior. By treating $f': x \rightarrow q(\tilde{X})$ as an equivalent model, we can leverage existing interpretability methods to generate explanations. We pick 75, 90, 95 quantiles of the output and get explanations correspondingly (Fig. ~\ref{fig:Probabilistic Adaption}).


\subsection{Setting Up Synthetic Benchmark}
\label{sec:datasets}

\changed{A key challenge inside the development of HTSF explainability methods is the lack of benchmarks due to the difficulty to determine and define ground truth explanations. To tackle this challenge, we establish} a synthetic benchmark with both known explanation ground truths and real-world distributions in three steps: first, we generate univariate synthetic series with a known ground truth explanation; second, we build hierarchical series with \rev{certain dependency relationships} and generate hierarchical sequence; third, we add the synthetic series to real-world datasets and perform evaluation based on output variations. 

\subsubsection{Univariate Synthetic Series Generation}\label{sec:UnivDataset}

We generate univariate synthetic series by adding anomalies to background Gaussian noise, \changed{which has been leveraged by previous univariate explanability benchmarks} ~\cite{turbe2023evaluation, queen2024encoding}. Specifically, while explaining a forecasting model, we expect the interpretability methods to assign higher importance scores on anomaly histories if the model is forecasting anomalies, or lower importance otherwise. Specifically, for anomalies with different mean, we expect the deterministic prediction results to assign higher importance; for anomalies with different variance, we expect the probabilistic prediction models to assign higher importance.

We introduced three types of anomalies: 

\begin{enumerate}
    \item Frequent shapes, where anomalies are composed of one-round sine waves with a certain frequency
    \item Sequence combination, where the anomalies contain high-frequency sine waves with average moving upwards or downwards across time.
    \item Low variance, where the anomalies are sampled from a Gaussian distribution whose variance is 1/10 of the rest \rev{of the} sequence.
\end{enumerate}

\subsubsection{Hierarchical Ground Truth Establishment}

Given the selected anomalies, we assign them into series in $G_{\tau}$ to generate hierarchical series with known explanations and correlations. Specifically, we consider three ways of building correlations:

\begin{enumerate}
    \item Same time-series, where each anomaly only exists in one specific variable. In this case, the correlation would only be between the variable and its parent.
    \item Cross time-series, where we assign each anomaly distribution to multiple variables who have the same parent, and thus create correlations between such variables.
    \item Cross level, where we assign each anomaly distribution to any groups of variables. This will bring us complex dependencies across such variables and their least-common-ancestor.
    \item External variables are assigned anomalous distributions along with time-series in the hierarchy. This will model dependencies between exogenous variables and variables in the hierarchy.
    \end{enumerate}

Besides the same-variable correlations, we also consider cross-variable correlations, where we adopt the above assignments between external variables and target variables.


\subsubsection{Add Synthetic Series to Real-world Datasets}

To better emulate the behavior of our method under real-world cases, we add the generated synthetic hierarchical series to real-world datasets. Such a dataset would contain both real-world distribution and known explanations. 
\changed{Instead of evaluating the interpretability methods' output directly, we calculate the variation of interpretability methods' output before/after adding synthetic patterns and evaluate such variations, so that the outputted explanation would solely align with ground truth from synthetic series and not be messed up with unknown internal explanations from the original data.} 

We added synthetic patterns to \changed{four} real-world hierarchical demand datasets, \changed{one from proprietary demand data from a large chemical company, and another three collected via public data sources}:

\begin{table}[h]
\centering
\begin{tabular}{lccccc}
\hline
Dataset & No. of Nodes & Levels of & Horizon & Obs. \\
\hline
Dow & 20022 & 8 & 24 & 1.2 M \\
M5 & 3914 & 12 & 12 & 7.7 M \\
Tourism-L   & 555 & 4,5 & 12 & 126k  \\
Wiki        & 207 & 5   & 1  & 696k  \\
\hline
\end{tabular}
\caption{\changed{Summary of demand datasets used in benchmark establishment. We covered a variety of node capacity, number of hierarchical levels, and horizon sizes in the selection of such datasets.}}
\label{tab:dataset_summary}
\end{table}

\begin{itemize}
    \item[1.] \textbf{Dow's Demand Forecasting}, which contains monthly historical sales of The Dow Chemical Company between January 2018 to June 2023, collected from 10+ industries across 160+ countries. There are 8 hierarchical levels in this dataset, going between basic products in certain local areas and the whole industry of a country. Besides historical sales, this dataset also contains some external business indicators that are chosen by domain experts. We treat sales history as the target variable and the business indicators as external variables.
    \item[2.] \textbf{M5 Dataset}, which is a public hierarchical forecasting dataset containing monthly retail sales of Walmart, aggregated per product, per sale center, and per area. This dataset contains 3914 series distributed across 12 hierarchical levels. We treat the sale history as the target variable. 
    \item[3.] \textbf{Tourism-L}~\cite{wickramasuriya2019optimal} contains time-series of the flow of tourism in different regions of Australia. The dataset contains two parallel hierarchies first grouped into regions (four levels) and second into mode of travel (five levels. The lower-level nodes are segregated on the basis of demographics.
    \item[4.] \textbf{Wiki}~\cite{taieb2017coherent} contains time-series of views of 145000 Wikipedia articles aggregated into 150 groups. These are further aggregated based on topic similarity into four levels.
\end{itemize}



\begin{table*}[ht]
\centering
\begin{tabular}{l|cccccccc|rrrrrrrr}
\toprule
Dataset & \multicolumn{8}{c|}{Dow's Demand} & \multicolumn{8}{c}{M5} \\ \midrule
Metric & \multicolumn{4}{c|}{IAS} & \multicolumn{4}{c|}{EVDA} & \multicolumn{4}{c|}{IAS} & \multicolumn{4}{c}{EVDA} \\ \midrule
\multicolumn{1}{c|}{Int. Method} & LIME & IG & FO & \multicolumn{1}{c|}{SG} & LIME & IG & FO & SG & \multicolumn{1}{c}{LIME} & \multicolumn{1}{c}{IG} & \multicolumn{1}{c}{FO} & \multicolumn{1}{c|}{SG} & \multicolumn{1}{c}{LIME} & \multicolumn{1}{c}{IG} & \multicolumn{1}{c}{FO} & \multicolumn{1}{c}{SG} \\ \midrule
FreqShapes & $0.39$ & $0.42$ & $0.37$ & \multicolumn{1}{c|}{$0.22$} & $0.38$ & $0.44$ & $0.48$ & $0.27$ & $0.56$ & $0.62$ & $0.37$ & \multicolumn{1}{r|}{$0.77$} & $0.52$ & $0.62$ & $0.54$ & $0.77$ \\
\multicolumn{1}{r|}{- w/o subtree} & $0.11$ & $0.17$ & $0.23$ & \multicolumn{1}{c|}{$0.17$} & $0.16$ & $0.32$ & $0.39$ & $0.28$ & $0.24$ & $0.36$ & $0.26$ & \multicolumn{1}{r|}{$0.49$} & $0.21$ & $0.37$ & $0.32$ & $0.46$ \\ \hline
SeqComb & $0.39$ & $0.33$ & $0.38$ & \multicolumn{1}{c|}{$0.29$} & $0.39$ & $0.42$ & $0.44$ & $0.32$ & 0.47 & 0.54 & 0.33 & \multicolumn{1}{r|}{0.73} & 0.58 & 0.74 & 0.37 & 0.64 \\
\multicolumn{1}{r|}{- w/o subtree} & $0.27$ & $0.24$ & $0.35$ & \multicolumn{1}{c|}{$0.28$} & $0.21$ & $0.24$ & $0.31$ & $0.29$ & 0.22 & 0.39 & 0.18 & \multicolumn{1}{r|}{0.53} & 0.33 & 0.31 & 0.24 & 0.35 \\ \hline
LowVar & $0.39$ & $0.32$ & $0.39$ & \multicolumn{1}{c|}{$0.38$} & $0.25$ & $0.42$ & $0.39$ & $0.22$ & 0.51 & 0.61 & 0.36 & \multicolumn{1}{r|}{0.74} & 0.66 & 0.65 & 0.42 & 0.59 \\
\multicolumn{1}{r|}{- w/o subtree} & $0.29$ & $0.22$ & $0.25$ & \multicolumn{1}{c|}{$0.31$} & $0.27$ & $0.38$ & $0.27$ & $0.17$ & 0.28 & 0.28 & 0.19 & \multicolumn{1}{r|}{0.44} & 0.37 & 0.26 & 0.21 & 0.33 \\ \bottomrule
\end{tabular}
\caption{Cross-variable deterministic hierarchical evaluation results on Dow's Demand and M5, averaged across all HTSF models. Subtree approximation improved 34 of 36 Dow experiments with relatively $40.4\%$ in EVDA and $62.0\%$ in IAS, and improved all M5 experiment with relatively $92.5\%$ in EVDA and $78.0\%$ in IAS.}\label{tab:pointCrossVariable}
\end{table*}

\begin{table*}[ht]
\centering
\scalebox{1.1}{
\begin{tabular}{c|l|cccc|cccc|cccc} \toprule
\multicolumn{1}{l}{} & Correlation Type & \multicolumn{4}{c}{Same Series} & \multicolumn{4}{c}{Cross Series} & \multicolumn{4}{c}{Cross Levels} \\ \cline{2-14}
\multicolumn{1}{l}{} & Int. Method & LIME & IG & FO & SG & LIME & IG & FO & SG & LIME & IG & FO & SG \\ \midrule
\multirow{6}{*}{\begin{tabular}[c]{@{}c@{}}Dow's\\ Demand\end{tabular}} & FreqShapes & 0.32 & 0.35 & 0.37 & 0.31 & 0.15 & 0.17 & 0.19 & 0.18 & 0.18 & 0.19 & 0.17 & 0.11 \\
 & \multicolumn{1}{r|}{- w/o subtree} & 0.29 & 0.37 & 0.34 & 0.33 & 0.13 & 0.18 & 0.17 & 0.15 & 0.15 & 0.19 & 0.17 & 0.09 \\ \cline{2-14}
 & SeqComb & 0.35 & 0.34 & 0.39 & 0.39 & 0.13 & 0.16 & 0.22 & 0.24 & 0.16 & 0.18 & 0.24 & 0.25 \\
 & \multicolumn{1}{r|}{- w/o subtree} & 0.25 & 0.36 & 0.32 & 0.39 & 0.17 & 0.21 & 0.19 & 0.15 & 0.19 & 0.24 & 0.2 & 0.11 \\ \cline{2-14}
 & LowVar & 0.21 & 0.24 & 0.24 & 0.23 & 0.11 & 0.24 & 0.18 & 0.19 & 0.1 & 0.12 & 0.12 & 0.11 \\
 & \multicolumn{1}{r|}{- w/o subtree} & 0.18 & 0.23 & 0.21 & 0.19 & 0.12 & 0.21 & 0.16 & 0.15 & 0.07 & 0.13 & 0.12 & 0.09 \\ \midrule
\multirow{6}{*}{M5} & FreqShapes & 0.19 & 0.25 & 0.25 & 0.25 & 0.19 & 0.24 & 0.22 & 0.19 & 0.17 & 0.21 & 0.18 & 0.18 \\
 & \multicolumn{1}{r|}{- w/o subtree} & 0.17 & 0.26 & 0.22 & 0.22 & 0.16 & 0.22 & 0.16 & 0.18 & 0.12 & 0.17 & 0.18 & 0.14 \\ \cline{2-14}
 & SeqComb & 0.22 & 0.29 & 0.16 & 0.21 & 0.23 & 0.25 & 0.16 & 0.18 & 0.18 & 0.19 & 0.13 & 0.14 \\
 & \multicolumn{1}{r|}{- w/o subtree} & 0.23 & 0.28 & 0.18 & 0.19 & 0.21 & 0.24 & 0.17 & 0.18 & 0.16 & 0.23 & 0.18 & 0.16 \\ \cline{2-14}
 & LowVar & 0.24 & 0.35 & 0.23 & 0.2 & 0.19 & 0.26 & 0.18 & 0.17 & 0.17 & 0.21 & 0.17 & 0.16 \\
 & \multicolumn{1}{r|}{- w/o subtree} & 0.21 & 0.31 & 0.21 & 0.18 & 0.19 & 0.22 & 0.22 & 0.12 & 0.16 & 0.26 & 0.16 & 0.11 \\ \midrule
\multirow{6}{*}{Tourism-L} & FreqShapes & 0.74 & 0.64 & 0.87 & 0.57 & 0.66 & 0.78 & 0.66 & 0.64 & 0.59 & 0.77 & 0.68 & 0.63 \\
 & \multicolumn{1}{r|}{- w/o subtree} & 0.59 & 0.74 & 0.81 & 0.67 & 0.52 & 0.62 & 0.63 & 0.53 & 0.48 & 0.66 & 0.66 & 0.55 \\ \cline{2-14}
 & SeqComb & 0.57 & 0.84 & 0.65 & 0.69 & 0.55 & 0.73 & 0.62 & 0.62 & 0.49 & 0.71 & 0.6 & 0.6 \\
 & \multicolumn{1}{r|}{- w/o subtree} & 0.47 & 0.75 & 0.75 & 0.68 & 0.45 & 0.69 & 0.59 & 0.62 & 0.38 & 0.63 & 0.55 & 0.57 \\ \cline{2-14}
 & LowVar & 0.46 & 0.58 & 0.48 & 0.59 & 0.41 & 0.54 & 0.37 & 0.53 & 0.33 & 0.41 & 0.38 & 0.48 \\
 & \multicolumn{1}{r|}{- w/o subtree} & 0.39 & 0.65 & 0.49 & 0.48 & 0.33 & 0.62 & 0.43 & 0.44 & 0.31 & 0.54 & 0.44 & 0.46 \\ \midrule
\multirow{6}{*}{Wiki} & FreqShapes & 0.56 & 0.46 & 0.65 & 0.46 & 0.39 & 0.42 & 0.39 & 0.37 & 0.24 & 0.36 & 0.32 & 0.31 \\
 & \multicolumn{1}{r|}{- w/o subtree} & 0.44 & 0.47 & 0.56 & 0.45 & 0.32 & 0.43 & 0.35 & 0.32 & 0.27 & 0.34 & 0.37 & 0.29 \\ \cline{2-14}
 & SeqComb & 0.59 & 0.58 & 0.41 & 0.56 & 0.34 & 0.56 & 0.34 & 0.52 & 0.31 & 0.43 & 0.31 & 0.54 \\
 & \multicolumn{1}{r|}{- w/o subtree} & 0.46 & 0.52 & 0.39 & 0.56 & 0.37 & 0.46 & 0.35 & 0.45 & 0.25 & 0.34 & 0.39 & 0.41 \\ \cline{2-14}
 & LowVar & 0.39 & 0.49 & 0.39 & 0.47 & 0.34 & 0.37 & 0.33 & 0.45 & 0.29 & 0.31 & 0.25 & 0.28 \\
 & \multicolumn{1}{r|}{- w/o subtree} & 0.36 & 0.41 & 0.32 & 0.41 & 0.31 & 0.32 & 0.27 & 0.37 & 0.29 & 0.25 & 0.21 & 0.29 \\ \bottomrule
\end{tabular}
}
\caption{Same-variable evaluation results on Dow's Demand, M5, Tourism-L, and Wiki. Among all the 36 experiments in each dataset, subtree approximation improved 22 Dow experiments with average $12.3\%$ improvement, 24 M5 experiments with average $8.0\%$ improvement, 26 Tourism experiments with average $7.5\%$ improvement, and 26 Wiki experiments with average $10.7\%$ improvement.}\label{tab:pointSame}
\end{table*}

\begin{table*}[ht]
\centering

\scalebox{1.1}{
\begin{tabular}{c|l|cccc|cccc|cccc} \toprule
 & Correlation Type & \multicolumn{4}{c}{Same Series} & \multicolumn{4}{c}{Cross Series} & \multicolumn{4}{c}{Cross Levels} \\ \cline{2-14}
 & Int. Method & LIME & IG & FO & SG & LIME & IG & FO & SG & LIME & IG & FO & SG \\ \midrule
\multirow{6}{*}{\begin{tabular}[c]{@{}c@{}}Dow's\\ Demand\end{tabular}} & FreqShapes & 0.28 & 0.26 & 0.34 & 0.23 & 0.08 & 0.19 & 0.14 & 0.13 & 0.08 & 0.14 & 0.12 & 0.07 \\
 & \multicolumn{1}{r|}{- w/o subtree} & 0.22 & 0.22 & 0.27 & 0.17 & 0.07 & 0.14 & 0.10 & 0.09 & 0.06 & 0.14 & 0.08 & 0.07 \\ \cline{2-14}
 & SeqComb & 0.25 & 0.42 & 0.29 & 0.43 & 0.20 & 0.27 & 0.18 & 0.17 & 0.14 & 0.15 & 0.19 & 0.16 \\
 & \multicolumn{1}{r|}{- w/o subtree} & 0.18 & 0.33 & 0.24 & 0.28 & 0.16 & 0.22 & 0.16 & 0.16 & 0.13 & 0.14 & 0.17 & 0.11 \\ \cline{2-14}
 & LowVar & 0.16 & 0.26 & 0.21 & 0.29 & 0.13 & 0.22 & 0.16 & 0.13 & 0.11 & 0.06 & 0.16 & 0.13 \\
 & \multicolumn{1}{r|}{- w/o subtree} & 0.13 & 0.24 & 0.17 & 0.22 & 0.10 & 0.16 & 0.12 & 0.12 & 0.08 & 0.04 & 0.13 & 0.11 \\ \midrule
\multirow{6}{*}{M5} & FreqShapes & 0.43 & 0.57 & 0.53 & 0.34 & 0.30 & 0.23 & 0.27 & 0.36 & 0.22 & 0.25 & 0.34 & 0.20 \\
 & \multicolumn{1}{r|}{- w/o subtree} & 0.30 & 0.56 & 0.45 & 0.37 & 0.29 & 0.24 & 0.28 & 0.31 & 0.22 & 0.19 & 0.27 & 0.20 \\ \cline{2-14}
 & SeqComb & 0.46 & 0.48 & 0.50 & 0.57 & 0.37 & 0.30 & 0.50 & 0.34 & 0.34 & 0.21 & 0.42 & 0.18 \\
 & \multicolumn{1}{r|}{- w/o subtree} & 0.44 & 0.32 & 0.47 & 0.54 & 0.27 & 0.28 & 0.34 & 0.36 & 0.30 & 0.16 & 0.29 & 0.15 \\  \cline{2-14}
 & LowVar & 0.66 & 0.72 & 0.28 & 0.35 & 0.43 & 0.49 & 0.14 & 0.20 & 0.31 & 0.28 & 0.11 & 0.23 \\
 & \multicolumn{1}{r|}{- w/o subtree} & 0.46 & 0.58 & 0.26 & 0.28 & 0.34 & 0.32 & 0.13 & 0.21 & 0.28 & 0.42 & 0.14 & 0.26 \\ \midrule
\multirow{6}{*}{Tourism-L} & FreqShapes & 0.38 & 0.58 & 0.51 & 0.45 & 0.29 & 0.40 & 0.48 & 0.33 & 0.32 & 0.33 & 0.34 & 0.40 \\
 & \multicolumn{1}{r|}{- w/o subtree} & 0.33 & 0.45 & 0.53 & 0.39 & 0.31 & 0.35 & 0.39 & 0.29 & 0.24 & 0.35 & 0.34 & 0.35 \\ \cline{2-14}
 & SeqComb & 0.39 & 0.70 & 0.79 & 0.48 & 0.40 & 0.63 & 0.57 & 0.46 & 0.34 & 0.60 & 0.40 & 0.52 \\
 & \multicolumn{1}{r|}{- w/o subtree} & 0.38 & 0.66 & 0.53 & 0.41 & 0.33 & 0.47 & 0.38 & 0.31 & 0.27 & 0.59 & 0.41 & 0.35 \\ \cline{2-14}
 & LowVar & 0.31 & 0.53 & 0.39 & 0.53 & 0.24 & 0.40 & 0.27 & 0.32 & 0.18 & 0.42 & 0.25 & 0.29 \\
 & \multicolumn{1}{r|}{- w/o subtree} & 0.29 & 0.47 & 0.34 & 0.35 & 0.21 & 0.39 & 0.25 & 0.28 & 0.20 & 0.31 & 0.23 & 0.30 \\ \midrule
\multirow{6}{*}{Wiki} & FreqShapes & 0.31 & 0.28 & 0.38 & 0.28 & 0.16 & 0.25 & 0.23 & 0.21 & 0.15 & 0.23 & 0.21 & 0.13 \\
 & \multicolumn{1}{r|}{- w/o subtree} & 0.24 & 0.27 & 0.32 & 0.20 & 0.17 & 0.20 & 0.16 & 0.16 & 0.10 & 0.17 & 0.15 & 0.10 \\ \cline{2-14}
 & SeqComb & 0.47 & 0.50 & 0.37 & 0.49 & 0.19 & 0.50 & 0.30 & 0.39 & 0.20 & 0.32 & 0.27 & 0.38 \\
 & \multicolumn{1}{r|}{- w/o subtree} & 0.35 & 0.38 & 0.26 & 0.36 & 0.20 & 0.34 & 0.24 & 0.31 & 0.14 & 0.24 & 0.20 & 0.28 \\ \cline{2-14}
 & LowVar & 0.37 & 0.34 & 0.31 & 0.36 & 0.23 & 0.28 & 0.16 & 0.28 & 0.17 & 0.17 & 0.13 & 0.18 \\
 & \multicolumn{1}{r|}{- w/o subtree} & 0.26 & 0.29 & 0.21 & 0.27 & 0.24 & 0.27 & 0.16 & 0.23 & 0.16 & 0.15 & 0.14 & 0.15 \\ \bottomrule
\end{tabular}
}
\caption{Probabilistic hierarchical explanation results on Dow's Demand, M5, Tourism-L, and Wiki, averaged across all HTSF models and three quantiles. Of all 36 experiments on each dataset, subtree approximation improved 34 Dow experiments with an average $26.0\%$ improvement, 29 M5 experiments with an average $18.1\%$ improvement, 29 Tourism experiments with an average $16.5\%$ improvement, and 31 Wiki experiments with average $25.3\%$ improvement.}\label{tab:probResult}
\end{table*}

\begin{figure*}[ht]
    \centering
    \subfloat[Absolute IAS with Subtree and relative improvement on Dow's Demand]{\includegraphics[width = .5\linewidth]{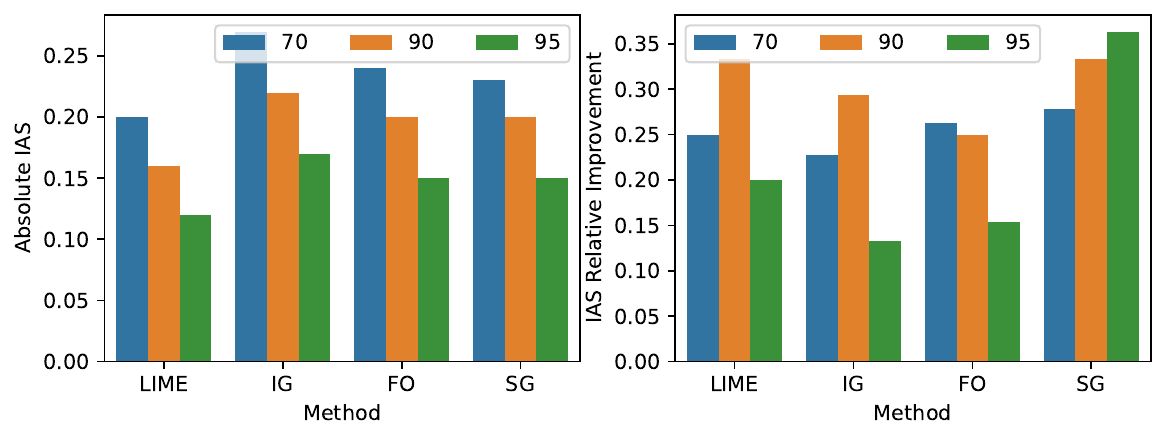}}
    \subfloat[Absolute IAS with Subtree and relative improvement on M5]{\includegraphics[width = .5\linewidth]{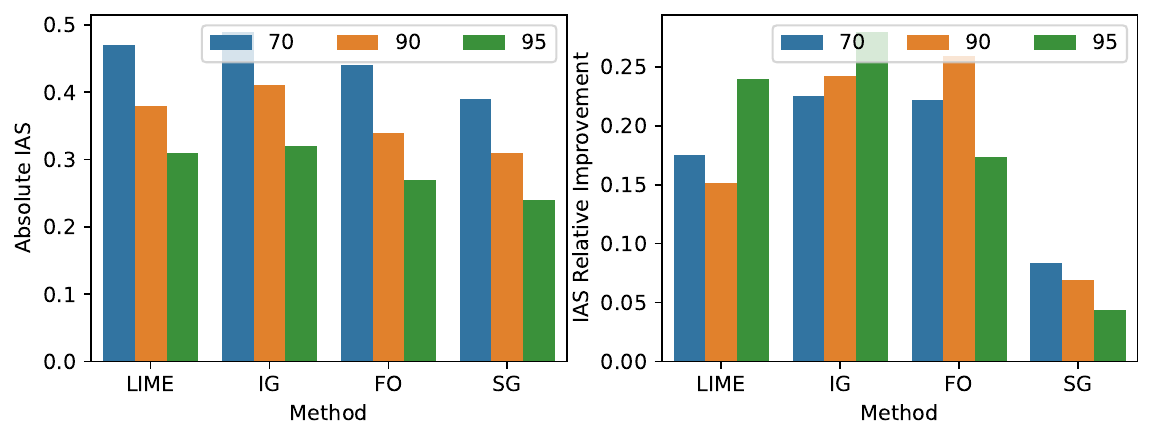}} \\
    \subfloat[Absolute IAS with Subtree and relative improvement on Tourism]{\includegraphics[width = .5\linewidth]{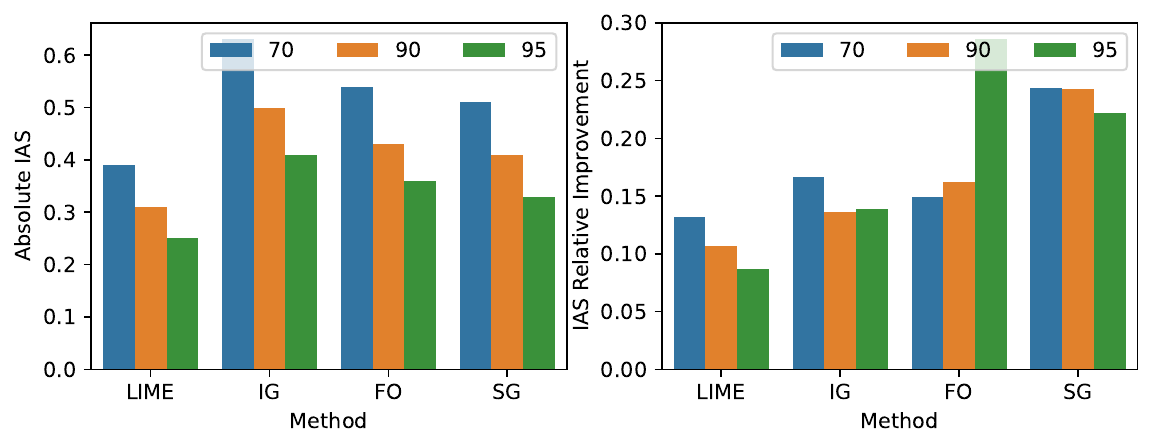}} 
    \subfloat[Absolute IAS with Subtree and relative improvement on Wiki]{\includegraphics[width = .5\linewidth]{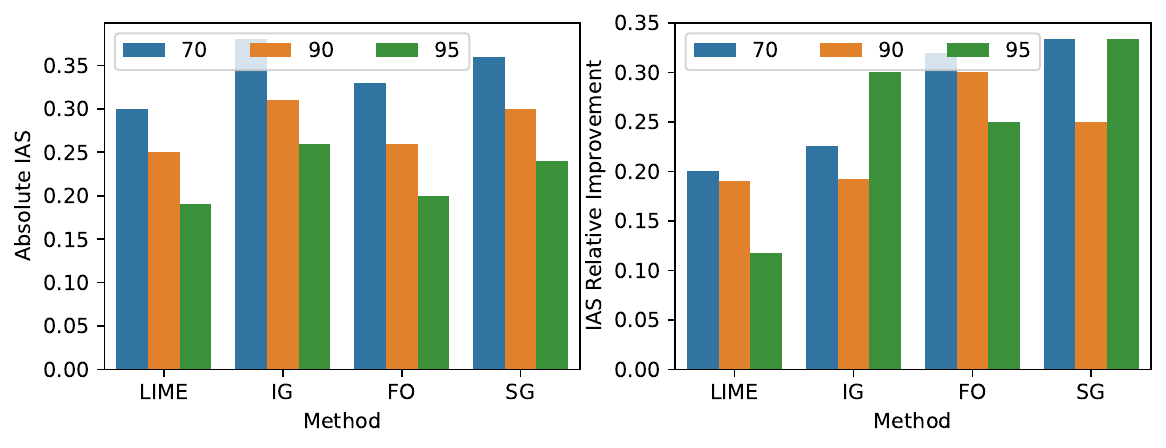}}
    \caption{Probabilistic evaluation results of each quantile. While the \rev{70th quantile} with subtree has the best absolute performance, the \rev{90th quantile} receives the most improvement by introducing subtree approximation. }
    \label{fig:quantileAblate}
\end{figure*}

\section{Experiments}

We evaluate \model over a wide range of datasets, benchmarks, and forecasting models. The code for implementation is provided at Github \footnote{\url{https://github.com/AdityaLab/Hiereinterpret}}.

\subsection{Settings}

For each synthetic dataset in \S \ref{sec:datasets}, we generate 1000 time-series datasets. We train the hierarchical forecasting model on the first 60\% of each of the time-series and produce forecasts for the rest of the
40\% of the time-series.
The explanations generated on the forecasts are compared against ground-truth explanations to evaluate the efficacy of our explanation method,
as well as other baselines on each of the forecasting models.

\subsubsection{Evaluation Metrics}
\paragraph{Importance Accuracy Score (IAS):}
The model output for each prediction is an importance score for each input value. For prediction $\hat{x}_{i}^t$,
let $I_{\hat{x}_{i}^t}(j, t')$ be the importance score of the input value $x_{j}^{t'}$.
We calculate the Importance Accuracy Score (IAS), which is a normalized importance score for time-series $j$ at time $t'$ regarding ground truth explanation $(x_{j^*}^{t^*})$, defined as
\begin{equation}
    \hat{I}_{\hat{x}_{i}^t}(j, t') = \frac{I_{\hat{x}_{i}^t}(j, t') - \frac{1}{T} \sum_k I_{\hat{x}_{i}^t}(j, k)}{\max_k I_{\hat{x}_{i}^t}(k, t') - \min_k I_{\hat{x}_{i}^t}(k, t')}.
\end{equation}

\paragraph{External Variable Detection Accuracy (EVDA):} In many cases, the stakeholders require information about which external variable was responsible for the outputted forecast. For a given forecast  $\hat{x}_{i}^t$,
we identify the most important external variable as
\begin{equation}
    E(\hat{x}_{i}^t) = \arg\max_l \max_k I_{\hat{x}_{i}^t}(l,k).
\end{equation}
i.e., the external variable $l$ with the highest score at any time stamp.
EVDA measures the fraction of times the explanation method identifies the correct external variable for all forecasts of a time-series dataset.

\begin{figure}[ht]
    \centering
    \subfloat[\rev{Avg. improvement for point forecasts}]{\includegraphics[width = .49\linewidth]{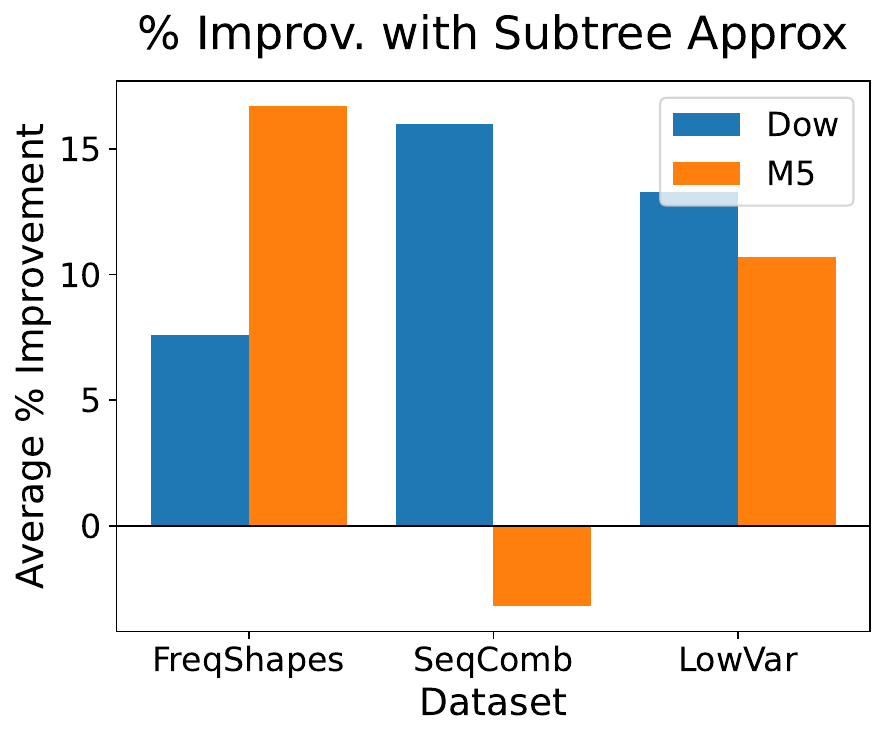}}\hfill
    \subfloat[\rev{Avg. improvement for probabilistic forecasts}]{\includegraphics[width = .49\linewidth]{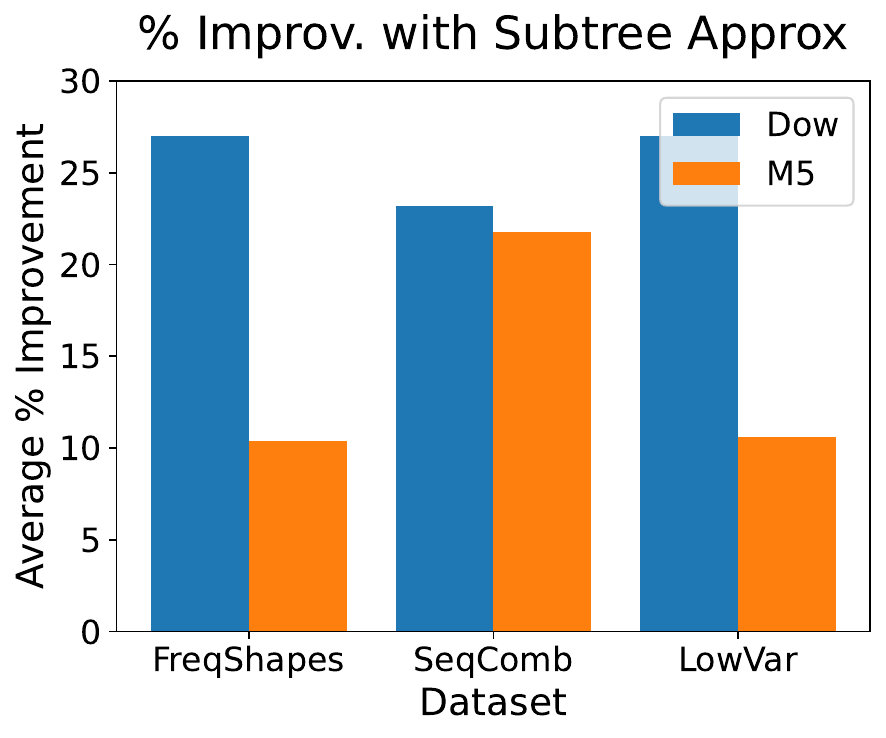}}
    \caption{Performance improvement of \model with Subtree approximation. Point forecasts see up to 16\% improvements, whereas probabilistic forecasts show over 25\% improvements}
    \label{fig:improvement}
\end{figure}

\subsubsection{Baseline Explanation Methods}

\begin{itemize}
    \item LIME (Local Interpretable Model-agnostic Explanations) \cite{ribeiro2016should}: LIME is a model-agnostic approximation-based explanation method that explains the predictions of any machine learning model by approximating it with a simpler model
    \item Feature Occlusion (FO) \cite{zeiler2014visualizing}: Feature occlusion is a perturbation-based method that measures the importance of each feature through changes in the model's prediction when the feature is occluded
    \item Integrated Gradients (IG) \cite{sundararajan2017axiomatic}: Integrated gradients is a gradient-based method that calculates importance by attributing the prediction to the input features by integrating the gradients of the model's prediction concerning the input features
    \item SmoothGrad (SG) \cite{smilkov2017smoothgrad}: SmoothGrad combines gradient-based and perturbation-based methods that assign importance by averaging the gradients over multiple noisy samples.
\end{itemize}

While LIME can be used for any machine learning model, the other methods are specific to neural networks.
The baseline forecasting models are implemented in Pytorch~\cite{paszke2019pytorch} and
 use captum~\cite{kokhlikyan2020captum} library to implement the interpretability methods
on all the models~\footnote{\url{https://captum.ai/}}.

\subsubsection{Interpreted HTSF Models}

\begin{figure*}[ht]
    \centering
    \subfloat[IAS scores of deterministic forecasting benchmark for different sizes of hierarchy]{\includegraphics[width = .5\linewidth]{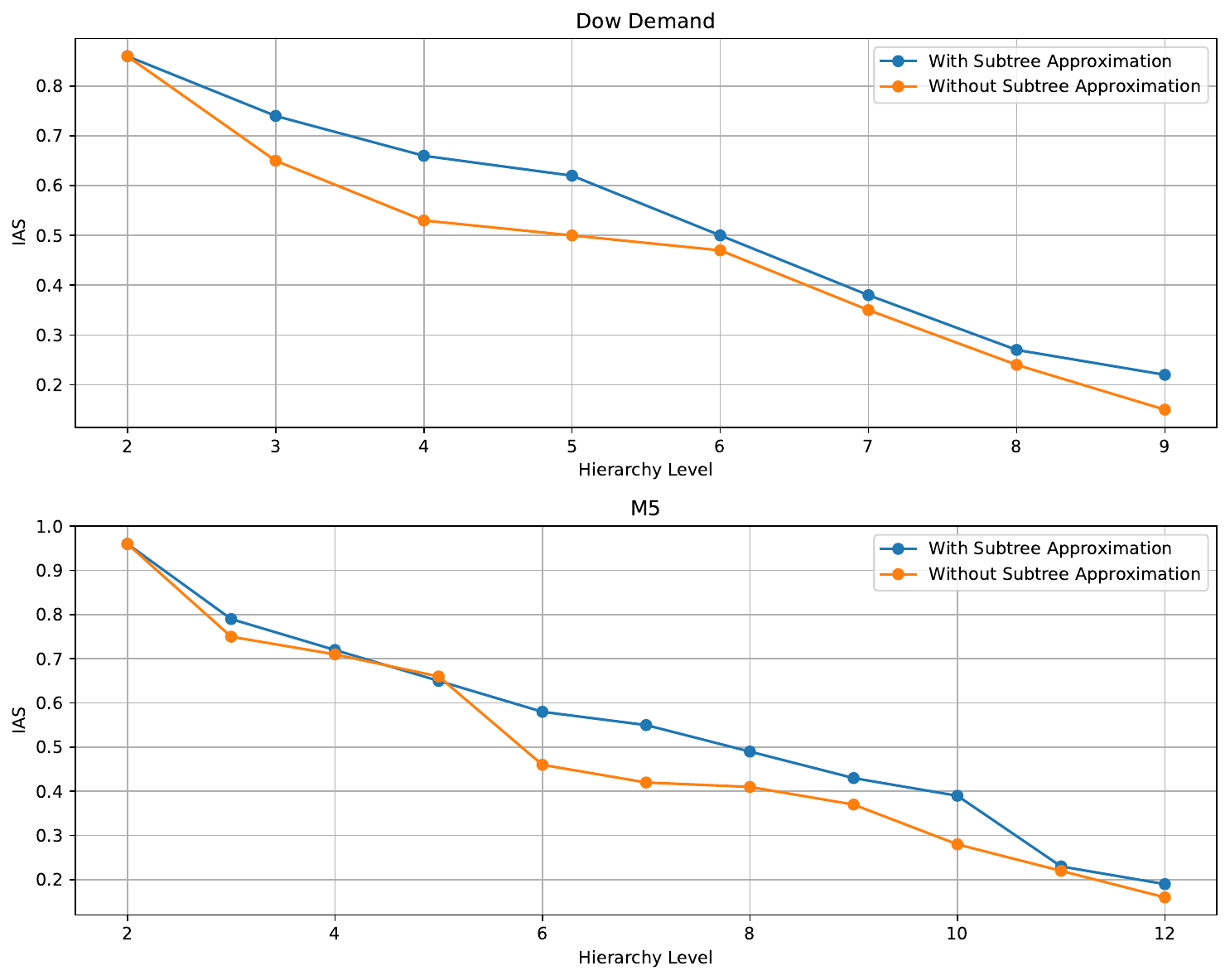}}
    \subfloat[IAS scores of probabilistic forecasting benchmark for different sizes of hierarchy]{\includegraphics[width = .5\linewidth]{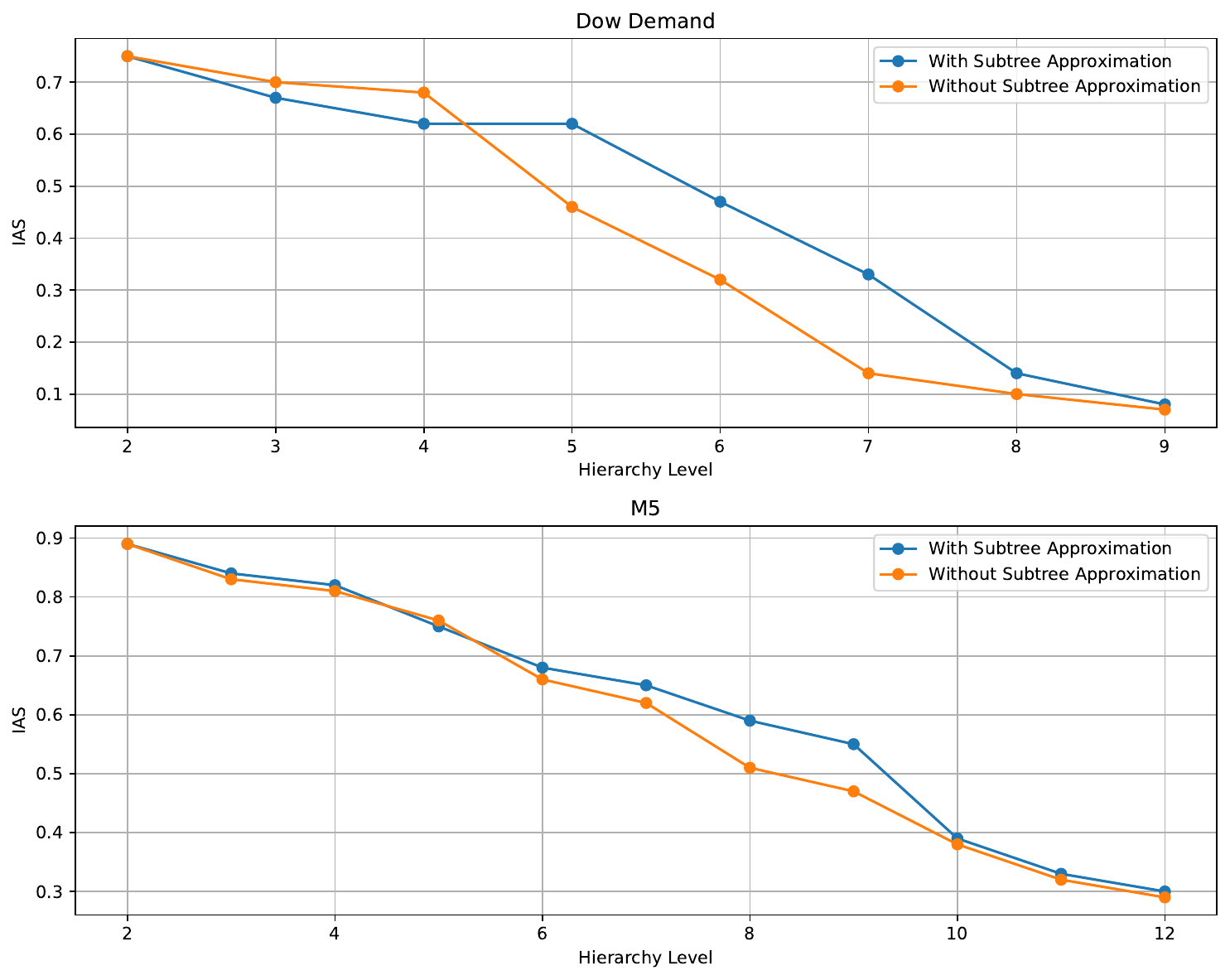}}
    \caption{Degradation of performance with increase in hierarchy size with and without subtree approximation}
    \label{fig:hierarchyimprov}
\end{figure*}

We introduce four state-of-the-art HTSF models and test our baseline explainability methods' performance on them.
\begin{itemize}
    \item HAILS \cite{kamarthi2024large}: HAILS is a hierarchical forecasting model that adapts to time-series of different sparsity by modeling their forecast distribution appropriately and reconciling them across the hierarchy.
    \item HierE2E \cite{rangapuram2021end}: HierE2E is a hierarchical forecasting model that uses a differentiable projection layer as the last layer, then projects the forecast to a subspace that adheres to the hierarchical relations.
    \item SHARQ \cite{han2021simultaneously}: SHARQ uses a quantile regression loss in the output of the base forecast that makes the values of each quantile consistent with the hierarchical structure.
    \item DeepAR \cite{salinas2020deepar}: DeepAR is a general-purpose forecasting model that uses a recurrent neural network to model the time-series and uses a likelihood loss to model the forecast distribution, typically parameterized by a normal distribution.
\end{itemize}

We use the default implementations from each of the method's authors without change in architecture and default hyperparameters for evaluation.
All the models are run on a single Nvidia V100 GPU with 32GB VRAM and Intel Xeon 80 Core CPU.

\subsubsection{Model Training}

To ablate the contribution of our method for explaining hierarchical and probabilistic models, we train HTSF models under two sets of settings: 

\begin{itemize}
    \item \textbf{Deterministic hierarchical forecasting}, where we treat the dataset as a deterministic series and optimize the model with mean squared error (MSE). 
    \item \textbf{Probabilistic hierarchical forecasting}, where we treat the dataset as samples from an unknown distribution and optimize the model with quantile regression loss as mentioned in Section \ref{sec:probAdapt}. 
\end{itemize}

In all settings, HTSF models are tuned to predict the next 12 months using the previous histories containing at least 12 histories.

\subsection{Results}

\rev{In summary, our experimental results demonstrate robust improvements of the proposed subtree approximation. In deterministic settings, our method outperformed the baselines in cross-variable explanations in 34 out of 36 experiments (Table \ref{tab:pointCrossVariable}), while simultaneously improving approximately $67\%$ of same-variable explanations (Table \ref{tab:pointSame}). Results from probabilistic experiments were equally compelling, with our approach showing improvement in roughly $80\%$ of cases (Table \ref{tab:probResult}), among which the 90th quantile yields the most improvements (Figure \ref{fig:quantileAblate}). Furthermore, our scalability (Figure \ref{fig:hierarchyimprov}) and computational complexity (Figure \ref{fig:runtime}) analyses confirm that the proposed method maintains high efficiency and accuracy across varying data scales. Additionally, we provide case studies on Dow's Demand forecasting and cross-verify our explanations with domain expert knowledge.}

\begin{figure*}[ht!]
    \centering
    \subfloat[Overview of the data. After the 22nd month (late 2019), Dow has a customer change which leads to the disappearance of huge periodic peaks.]{\includegraphics[width =0.5\linewidth]{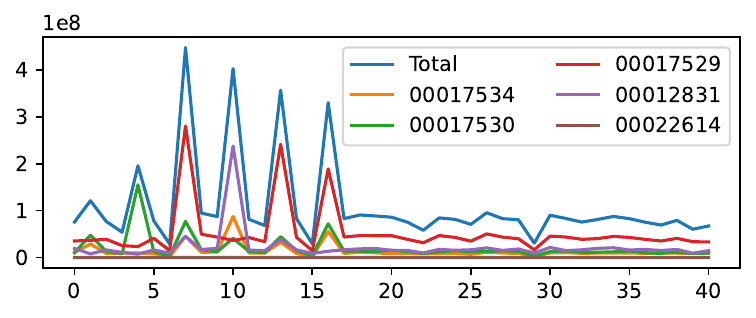}}\label{fig:VarianceCSOverview}
    \hfill
    \subfloat[Prediction of HAILS and assigned importance by \model on certain time-step (red square), before accessing new data.]{\includegraphics[width = 0.24\linewidth]{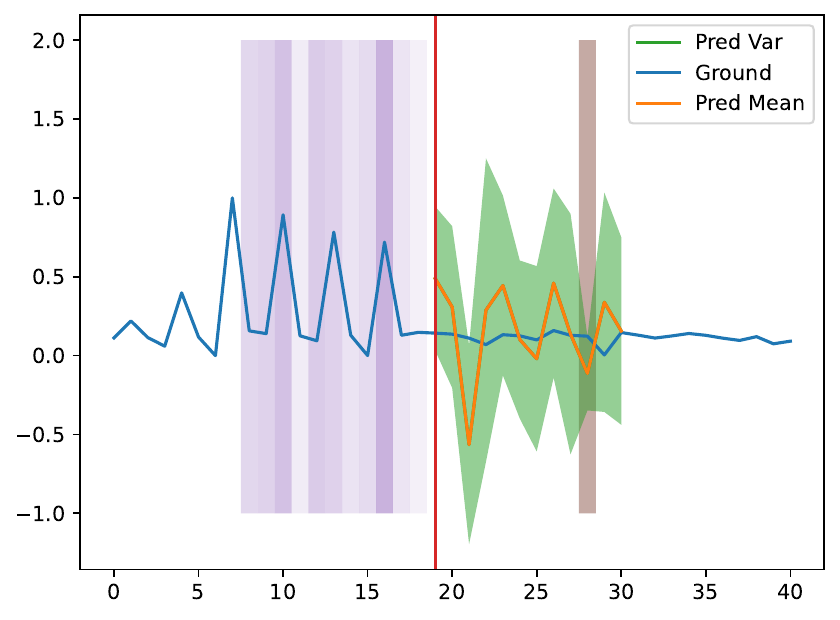}}\label{fig:VarianceCSBefore}
    \hfill
    \subfloat[Prediction of HAILS and assigned importance by \model on certain time-step (red square), after accessing new data.]{\includegraphics[width = 0.24\linewidth]{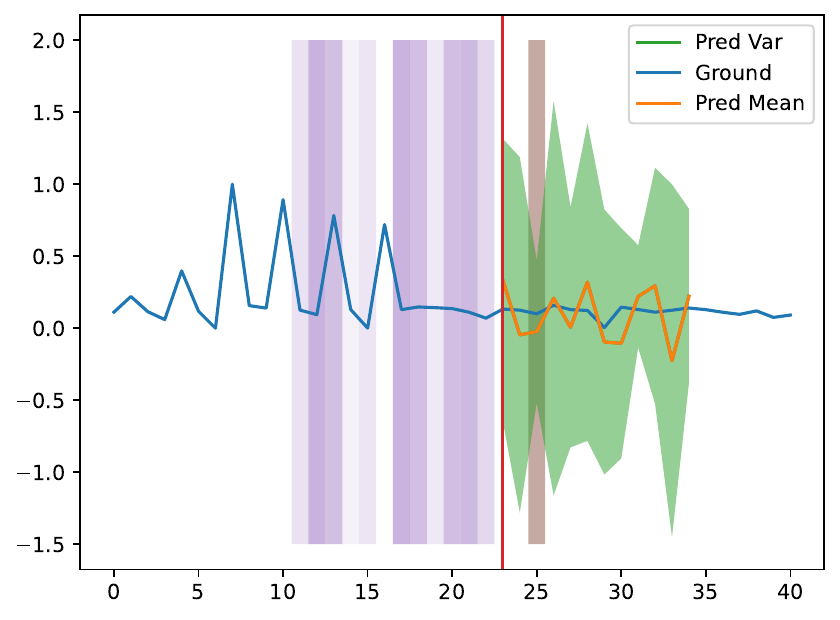}}\label{fig:VarianceCSAfter}
    
    \caption{After accessing new data without periodic spikes, HAILS tend to give high uncertainty in its prediction. \model shows that such high uncertainty output comes from the focus on two types of data -- with and without periodic spikes.}
    \label{fig:enter-label}
\end{figure*}

\subsubsection{RQ1: Deterministic Hierarchical Prediction}

We assign anomalies to all variables on the synthetic dataset established on Dow's dataset, train HTSF models on the generated dataset, and run interpretability methods on the trained models.
Table \ref{tab:pointCrossVariable} shows the results averaged on all four HTSF models. Our subtree approximation improved 34 of the 36 total experiments with an average $40.4\%$ improvement in EVDA and $62.0\%$ improvement in IAS, confirming that our approximation helps reduce the effect of mid-way noises and improve the explanation quality greatly. Also, among all the interpretability methods, Integrated Gradient outperforms the rest on most synthetic datasets while LIME and Feature Occlusion follow, showing that gradient-based methods are the best choice to explain cross-variable dependency in black-box HTSF models.

Furthermore, we assign anomalies to the target variable according to the three aforementioned hierarchical relationships and evaluate \model. 
As shown in Table \ref{tab:pointSame}, our subtree approximation improved 22 of 36 Dow experiments with an average $12.3\%$ relative improvement, 24 of 36 M5 experiments with an average $8.0\%$ relative improvement, 26 of 36 Tourism experiments with average $7.5\%$ relative improvement, and 26 of 36 Wiki experiments with average $10.7\%$ relative improvement. Generally, we see that the absolute IAS values and subtree improvements are not as significant as all-variable results, implying that cross-hierarchy dependencies are more difficult to explain than cross-variable dependencies. 


\begin{figure}[ht] 
    \centering
    \subfloat[Forecasts and importance score before beginning of periodic upward trends]{\includegraphics[width=0.45\linewidth]{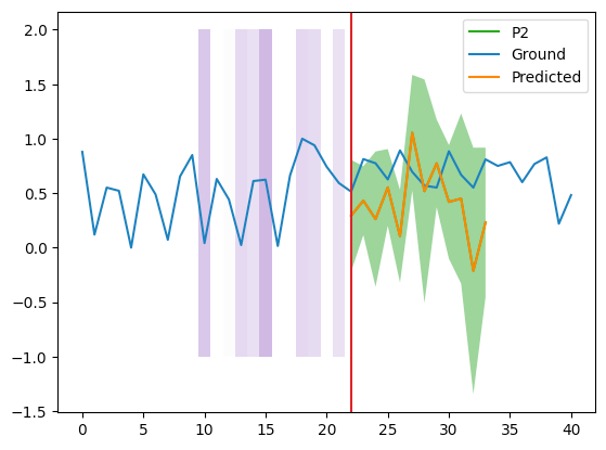}}
        \label{fig:Periodic1}
    \hfill
    \subfloat[Forecasts and importance scores after beginning of periodic upward trends]{\includegraphics[width=0.45\linewidth]{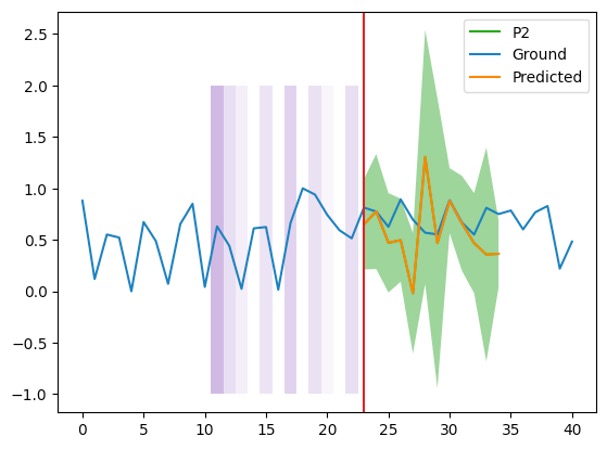}}
        \label{fig:Periodic2}
\caption{As the model anticipates the beginning of the periodic trends, \model notes the increased importance of the periodic trends from the past.}
\label{fig:periodicCD}
\end{figure}

\subsubsection{RQ2: Probabilistic Hierarchical Prediction}

We evaluated the quantile explanations following the same settings in deterministic same-variable predictions. As shown in Table \ref{tab:probResult}, our subtree approximation improved the performance on 34 of 36 Dow experiments with an average $26.0\%$ improvement, 29 of 36 M5 experiments with an average $18.1\%$ improvement, 29 of 36 Tourism experiments with an average $16.5\%$ improvement, and 31 of 36 Wiki experiments with an average $25.3\%$ improvement. Furthermore, Figure \ref{fig:quantileAblate} shows the absolute IAS with subtree approximation and relative improvement by introducing subtree. We see \rev{70th quantile} explanations match the ground truth most with an average of $0.21$ IAS, which is not surprising as such a quantile is frequently used in statistical summaries to represent a reasonable border of a distribution. Surprisingly, we see the improvement of the subtree on the \rev{90th quantiles} is close to that on \rev{70th quantiles}, followed closely by 95 quantiles. This implies that our subtree approximation brings similar improvement to all scenarios, regardless of previous performance.

\subsubsection{RQ3: Scalability with size of the hierarchy}
One of the main challenges of performing interpretability analysis over hierarchical time-series is the handling of a large number of time-series across different levels of the hierarchy.
We run an ablation study on the Dow and M5 datasets by considering the time-series
up to specific levels of the hierarchy, and measuring the degradation of 
performance with an increase in the number of hierarchy levels.

We evaluate the IAS scores for different levels of hierarchy for both datasets in Fig. ~\ref{fig:hierarchyimprov}.
We observe that for most levels of hierarchy, using subtree approximation yields better
accuracy of explanations for both deterministic and probabilistic forecasting settings.
This shows that sub-tree approximation is applicable to hierarchical forecasting scenarios of different scales.

\rev{
\subsubsection{RQ4: Running time and algorithmic complexity}
Our method is designed with accuracy and efficiency as primary goals. Applying any popular baseline without leveraging the hierarchical structure by considering all possible nodes of the tree for forecast from every node will yield complexity of $O(N^2 K_N)$ where $N$ is the number of nodes of the hierarchy and $K_N$ is the complexity of running the interpretable method on $N$ variables. Our algorithm essentially uses BFS traversal to assign the scores across the hierarchy altogether with runtime complexity $O(N K_T(i))$ for each node $i$. There $T(i)$ is the number of children of node $i$. The total complexity is $O(N^2 K_{T_{max}})$ where $T_{max}$ is the maximum number of children any node has in the hierarchy. Ignoring corner cases like a star graph, $T_{max} << N$ in most cases.
}

\begin{figure}[h]
    \centering
    \includegraphics[width=\linewidth]{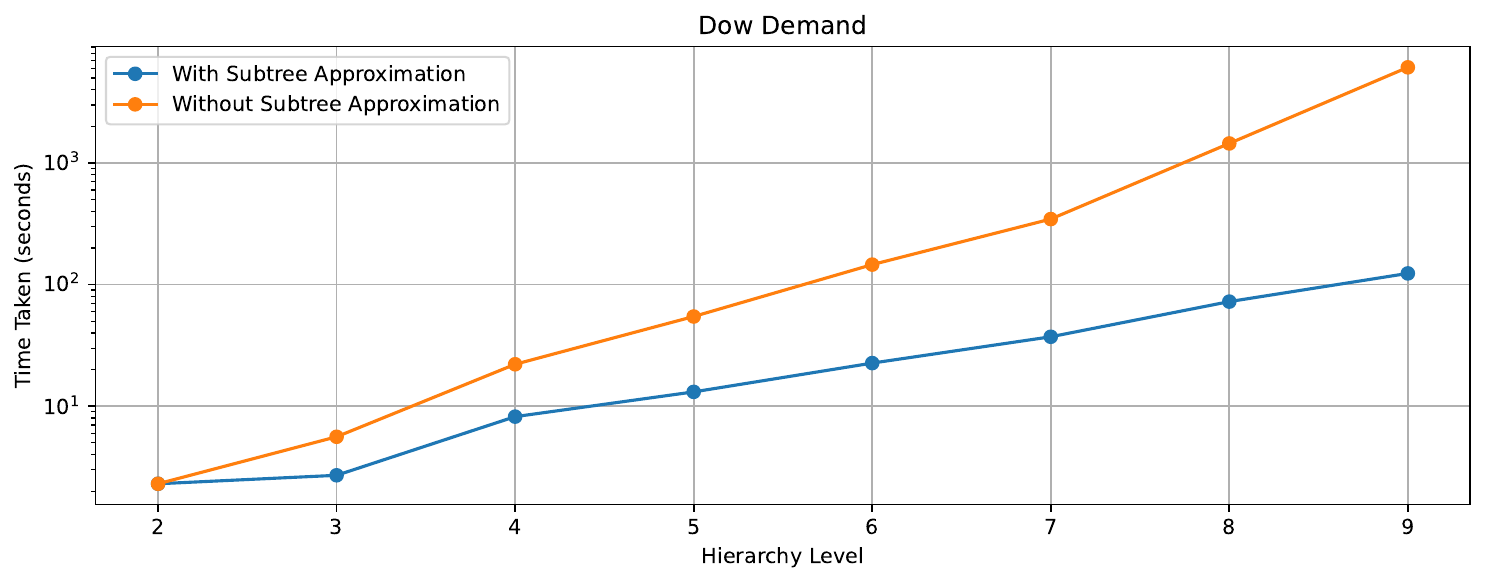}
    \caption{\rev{Running time of explanation models on HAILS considering hierarchies up to different levels. Our proposed subtree approximation greatly cuts down the time complexity.}}
    \label{fig:runtime}
\end{figure}

\rev{
We also designed experiments to measure the run-time with increasing hierarchical levels for the Dow dataset with IG in Fig.~\ref{fig:runtime}.
We ran the experiments on a server with Tesla V100 (32GB) GPUs with Intel Xeon(R) E5-2698 80-core CPU and 512GB of memory.
We run the explainability algorithm on HAILS during inference.
We see that the runtime improvement with subtree approximations increases significantly with an increase in the levels of the hierarchy. In fact, running the algorithm on the full tree takes 123.5 seconds using sub-tree approximation compared to 6114.8 seconds using all variables for each node independently.
}

\subsection{Case Study}

In addition to previous results from synthetic benchmarks, we present case studies of how our model performed on purely real-world series. We pick the sales history of all products from one specific value center from Dow's Demand and train HAILS \cite{kamarthi2024large}\changed{, one HTFS model that has achieved state-of-the-art performance on Dow's Demand dataset}, following the same setting as the main experiments.

\subsubsection{\model identify important trends from the past that are used by \rev{the} forecasting model}
We pick the demand history of some housing-related products between 2018 and 2023. This data has strong periodicity but has shifted upwards after late 2019 (22nd month). Dow's domain experts explain that this is caused by the COVID pandemic: as people tend to stay home during the pandemic, their willingness to furnish their house increases, and the corresponding demand grows. We limit the accessible time range of data before and after the 22nd month, get predictions from HAILS, and apply \model to explain the predictions. As shown in Fig.\ref{fig:Periodic1}, before accessing COVID data, \model provides more importance to downward parts of the past (Fig. \ref{fig:Periodic1}). However, once the model anticipates an upward trend in the future 
based on historical data, it predicts a future upward trend with higher importance scores to the time-steps with this trend in the past (Fig. \ref{fig:Periodic2}). This aligns with our domain knowledge, confirming that \model captures the model's focus under changing inputs well.

\subsubsection{Detection of forecast changes due to key economic indicators}
\begin{figure}[h!] 
    \centering
    \subfloat[Demand forecast before change in trend]{\includegraphics[width=0.45\linewidth]{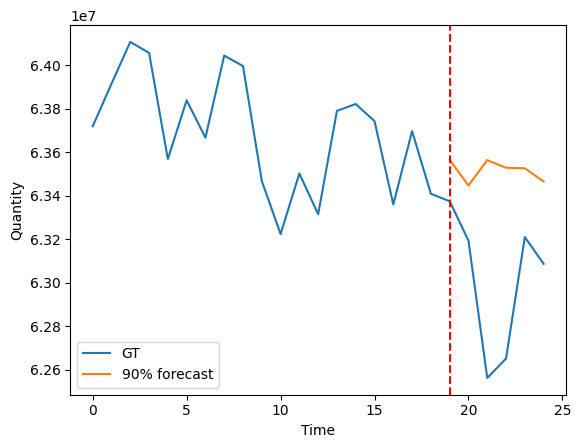}}
        \label{fig:Pred1}
    \hfill
    \subfloat[Demand forecast after change in trend]{\includegraphics[width=0.45\linewidth]{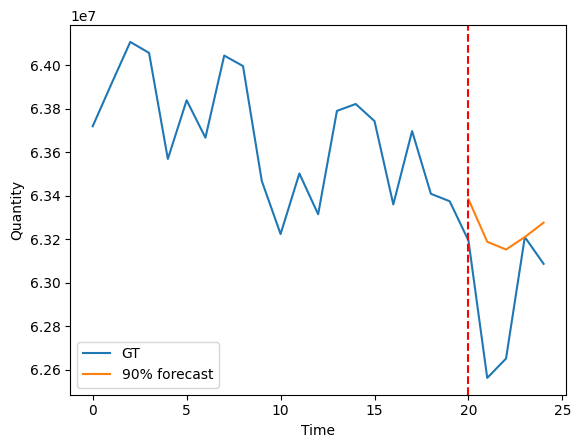}}
        \label{fig:Pred2}
    \\
    \subfloat[Importance of the economic indicator before change in trend]{\includegraphics[width=0.45\linewidth]{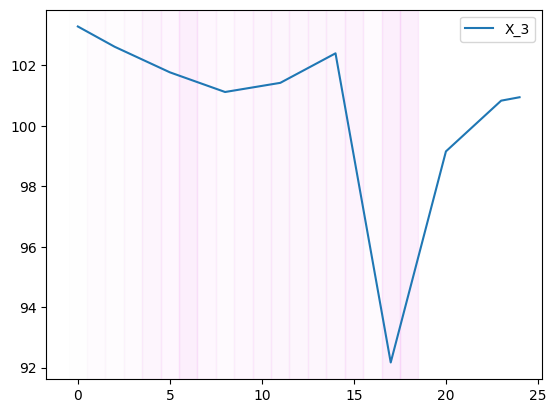}}
        \label{fig:Econ1}
    \hfill
    \subfloat[Importance of the economic indicator after change in trend]{\includegraphics[width=0.45\linewidth]{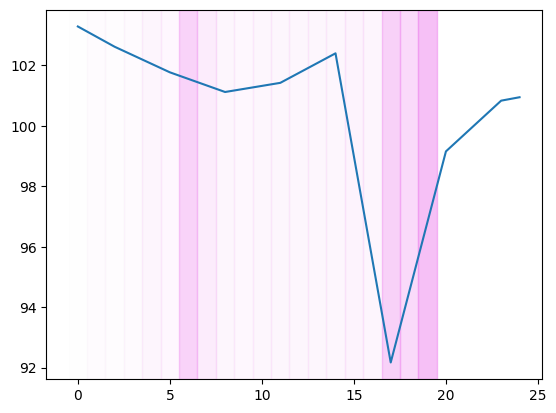}}
        \label{fig:Econ2}
    \caption{The downward spike observed in the second forecast causes \model to interpret that the important economic variable's previous downward trend may have a significant impact on the change in forecast trend.}
    \label{fig:UncertaintyCS}
\end{figure}

\model allows us to identify important variables of interest across the hierarchy as well
as across many external variables.
As shown in Fig.\ref{fig:UncertaintyCS}, we applied \model to explain the prediction of HAILS on the demand history of some packaging-related products between 2018 and 2021.
The 90\% forecasts of HAILS changed before (Fig.~\ref{fig:Pred1}) and after (Fig.~\ref{fig:Pred2}) observing the start of a downward trend around late 2019 (17th month). \model claims that such change is caused by a sudden change of a key economic indicator, Consumer Price Index (CPI), as shown in Fig. \ref{fig:Econ1} and Fig. \ref{fig:Econ2}. This explanation is agreed by Dow's domain expert: As CPI once goes down, people lose interest in purchasing, which makes packaging demand go down temporarily.
This shows that \model can identify useful relations across time-series from economic variables to demand forecasts across the hierarchy. 

\subsubsection{Changed Variance after Customer changes}

To show how the uncertainty explanation of \model performs, we pick the demand history of some heating products between 2018 and 2021, consisting of 5 histories from different centers. As shown in Fig.\ref{fig:VarianceCSOverview}, Dow terminated their collaboration with one major customer in early 2020, which made the following series lose periodic peaks. We set two limitations on data accessibility, before and after the aforementioned event occurred, applied HAILS, and explained the predictions by \model. We see when HAILS only has older data available (Fig.\ref{fig:VarianceCSBefore}), it gives periodic predictions with low uncertainty, and \model explanation focuses on previous periodic peaks. Yet, after accessing new data, HAILS give low predictions with high uncertainty, as shown in Fig.\ref{fig:VarianceCSAfter}. \model explains that, as HAILS accesses two data patterns, it puts equal concentration on both and therefore becomes uncertain of its prediction.

\section{Conclusion}

In this paper, we studied the important problem of providing accurate explanations for large-scale hierarchical time-series forecasting.
This problem was studied in the context of deploying reliable demand forecasting
for thousands of products at a large chemical company, identifying key challenges
common in many industrial applications.
We tackled the challenge of handling thousands of time-series in the hierarchy robustly
to identify accurate explanations across the hierarchy as well as external variables,
including accuracy, efficiency, and adaptation to probabilistic forecasts.

We proposed two improvements to help adapt the interpretability methods designed for general machine learning models into HTSF tasks.
Subtree approximation helps interpretability models handle variables across the hierarchy efficiently and effectively.
Non-parameterized translation helps the model provide explanations to arbitrary forecast distributions
by leveraging quantile-specific importance scores for probabilistic explanations.
Experiments on the benchmarks generated from a wide range of datasets, including the real-world Dow industrial dataset, show that our methods, both subtree approximation and quantile adaptation, help improve the explanation accuracy by 12.3\% and 26\% for deterministic and probabilistic forecasting, respectively.
We also show that our method better scales with increasing hierarchy size.
We also provided multiple case studies on Dow's Demand dataset that show our method identifies important trends from the past, key economic indicators that caused significant shifts in predictions, as well as predicted useful explanations for changes in variance and uncertainty of time-series forecasts.

\rev{Furthermore, our advancements in interpretability open a wide range of future applications for demand forecasting. First, while our current approach is limited to numerical time-series data, it could be extended to incorporate exogenous information from multimodal inputs, such as geographical features, demographic documents, or mechanistic knowledge from domain experts. Second, our framework could facilitate the development of self-reflection systems that enhance the forecasting quality of existing models by generating feedback based on interpretability outputs. Finally, our model can be deployed across diverse business scenarios to provide transparency and support for otherwise "black-box" decision-making models.}

\section{Acknowledgements}

This paper was supported in part by The Dow Chemical Company, the NSF
(Expeditions CCF-1918770, CAREER IIS-2028586, Medium IIS-1955883, Medium IIS-2403240, Medium IIS-2106961), NIH (1R01HL184139), CDC MInD program, Meta, and Dolby faculty gifts.

\bibliographystyle{IEEEtran}
\bibliography{main}

\end{document}